\documentclass[10pt,journal,compsoc]{IEEEtran}
\ifCLASSOPTIONcompsoc
  \usepackage[nocompress]{cite}
\else
  \usepackage{cite}
\fi

\ifCLASSINFOpdf
\else
\fi

\usepackage{graphicx}
\usepackage{epsfig}
\usepackage{amsmath}
\usepackage{mathtools}
\usepackage{xspace}
\usepackage{algorithm}
\usepackage{algorithmic}
\usepackage{amssymb}
\usepackage{bm}
\usepackage{bbding}
\usepackage{booktabs} 
\usepackage{tabularx}
\usepackage{multirow}
\usepackage{etoolbox,siunitx}
\usepackage{color}      
\usepackage{diagbox}
\usepackage[font=scriptsize,skip=2pt,subrefformat=parens]{subcaption}
\usepackage[colorlinks=true, allcolors=blue]{hyperref}
\usepackage{ragged2e}
\usepackage{float}
\usepackage{bbding}
\usepackage{subcaption}
\usepackage{makecell}

\newcommand{\etal}{\textit{et al}. }
\newcommand{\ie}{\textit{i}.\textit{e}.}
\newcommand{\eg}{\textit{e}.\textit{g}.}
\newcommand{\etc}{\textit{etc}. }

\begin{document}
%
\title{
    Modeling Weather Uncertainty for \\ Multi-weather Co-Presence Estimation
}

\author{Qi Bi~\IEEEmembership{Student~Member,~IEEE}, Shaodi You,~\IEEEmembership{Senior~Member,~IEEE}, Theo Gevers~\IEEEmembership{Member,~IEEE}}

\markboth{Journal of \LaTeX\ Class Files,~Vol.~14, No.~8, August~2015}%
{Shell \MakeLowercase{\textit{et al.}}: Bare Demo of IEEEtran.cls for Computer Society Journals}
%

\IEEEtitleabstractindextext{%
\justify
\begin{abstract}
Images from outdoor scenes may be taken under various weather conditions. It is well studied that weather impacts the performance of computer vision algorithms and needs to be handled properly. However, existing algorithms model weather condition as a discrete status (\eg, sunny, cloudy, rainy and foggy) and estimate it using multi-label classification.
The fact is that, physically, specifically in meteorology, weather are modeled as a continuous and transitional status (e.g. sunny, cloudy, rainy and foggy are defined depending on the presence of water).
Instead of directly implementing hard classification as existing multi-weather classification methods do, we consider the physical formulation of multi-weather conditions and model the impact of physical-related parameter on learning from the image appearance.
In this paper, we start with solid revisit of the physics definition of weather and how it can be described as a continuous machine learning and computer vision task. Namely, we propose to model the weather uncertainty, where the level of probability and co-existence of multiple weather conditions are both considered.
A Gaussian mixture model is used to encapsulate the weather uncertainty and a uncertainty-aware multi-weather learning scheme is proposed based on prior-posterior learning. 
A novel multi-weather co-presence estimation transformer (MeFormer) is proposed.
In addition, a new multi-weather co-presence estimation (MePe) dataset, along with 14 fine-grained weather categories and 16,078 samples, is proposed to benchmark both conventional multi-label weather classification task and multi-weather co-presence estimation task.
Large scale experiments show that the proposed method achieves state-of-the-art performance and substantial generalization capabilities on both the conventional multi-label weather classification task and the proposed multi-weather co-presence estimation task.
Besides, modeling weather uncertainty also benefits adverse-weather semantic segmentation.
Ablation studies demonstrate the influence of each component of the proposed method. The new dataset and source code will be publicly available.
\end{abstract}

}

\maketitle

\IEEEdisplaynontitleabstractindextext
\IEEEpeerreviewmaketitle

\IEEEraisesectionheading{\section{Introduction}\label{intro}}

\IEEEPARstart{I}{mages} taken from outdoor scenes may contain various weather conditions such as sun, rain and fog \cite{chen2022learning,musat2021multi,sakaridis2021acdc}. These weather conditions can negatively influence the accuracy of outdoor computer vision algorithms such as object recognition, scene segmentation and 3D reconstruction \cite{choi2021robustnet,bi2023learning,diaz2022ithaca365,mirza2022efficient,Bi2023AD}. Therefore, existing methods focus on improving the visibility (\eg, dehazing \cite{zhou2021learning,li2017haze}, deraining \cite{liang2022drt,you2015adherent}) or adapting the scene representation to a weather-specific domain \cite{sakaridis2021acdc,dai2020curriculum,ma2022both,Bi2024Generalized}. Performing such works relies on a deterministic and known weather condition \cite{han2021joint}. 

So far, only a few methods are proposed to recognize weather conditions \cite{lin2017rscm,zhao2018cnn,xie2021graph,xie2022learning,xie2022wcatn}. Unfortunately, existing methods consider the weather condition in a binary and deterministic way, \ie, is it sunny? Yes/No. Is it rainy? Yes/No. Such binary predictions often do not correspond to reality.

\begin{figure}[!t]
  \centering
   \includegraphics[width=1.0\linewidth]{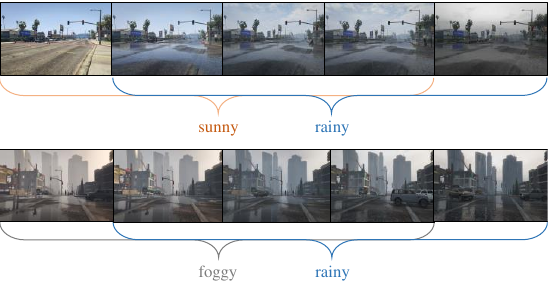}
   \vspace{-0.6cm}
   \caption{Discrete labels are insufficient to describe weather because it is transitional.
   An outdoor scene can observe the co-existence of different weather conditions under a variety of probability levels.
   \textbf{First row:} sunny and rainy; \textbf{Second row:} foggy and rainy.
   }
   \label{observation}
\end{figure}

From the meteorology view, weather is the state of the atmosphere in a continuous and transitional way. Which is impacted by multiple variables such as
moisture inflow, moisture outflow, evapotranspiration, precipitation and \etc \cite{zhang2004some}. Combination between multiple variables can lead to dramatically different appearance of a certain weather from the visual perception.
In fact, in most cases, the weather is a continuous condition with various levels of probabilities (see Fig.~\ref{observation} for an example), which makes the conventional binary and deterministic estimation less feasible. 
\textit{To the best of our knowledge}, no existing method has considered the weather as a continuous condition.

This paper provides a new perspective for multi-weather co-presence estimation. 
We systematically model the physical formulation of multi-weather conditions and analyze how the physical-related parameters impact the visual representation learning.
We consider each weather attribute in images taken from outdoor scenes as continuous. 
Besides, the weather conditions are considered with different levels of probability and co-occurrences, and are represented by a Gaussian mixture model \cite{rasmussen1999infinite,kendall2017uncertainties}.
To this end, an uncertainty-aware multi-weather learning scheme is proposed. 
In this way, the overlap between each weather representation is more flexible and realistic to depict the mixture of multi-weather conditions.

Then, a prior-posterior scheme is proposed \cite{ciosek2020conservative,Malinin2018Predictive} to explicitly learn the weather uncertainty. On the one hand, we need certain weather conditions to be clustered in feature space, and on the other hand, we need to represent the uncertainty of the weather features. For this, a prior-posterior learning pipeline is proposed. Finally, features are projected to probabilities of weather conditions. 

A new \textbf{M}ulti-weather Co-presence \textbf{E}stimation Trans\textbf{former} (MeFormer) is proposed. A set of \textit{weather tokens} are embedded into the Swim-Transformer \cite{liu2021swin} to generate the initial feature representation. The prior and posterior representations are learnt by two separate conditional variational auto-encoder (cVAE) \cite{sohn2015learning}. Probability prediction uses fully connected layers with activation, which can be seamlessly integrated into other down-stream tasks.
 
A new \textbf{m}ulti-w\textbf{e}ather Co-\textbf{p}resence \textbf{e}stimation (MePe) dataset is provided containing 16,078 samples of various outdoor scenes and 14 different weather conditions. We use the Grand Theft Auto V (GTAV) game engine to render the weather. In this way, we can steer and quantize the weather. 
To the best of our knowledge, it is so far the only multi-weather co-presence estimation dataset that can provide both binary labels and probability labels. Notably, the probability-level ground truth is tailored to describe the weather as a continuous condition and to support the multi-weather co-presence estimation task.
Four evaluation metrics, namely, the sum of squared difference (SSD), the Kullback-Leibler divergence (KL), coefficient of determination ($R^{2}$) and cross-entropy (CE), are given together with the dataset.

Large-scale experiments are conducted showing the performance of the proposed MeFormer and the MePe dataset. 
In addition, the proposed MeFormer also achieves the state-of-the-art multi-weather classification performance on Weather Attribute Dataset \cite{lin2017rscm}, Multi-weather Recognition Dataset \cite{zhao2018cnn} and the proposed MePe dataset.
Moreover, we also demonstrate that modeling weather uncertainty can substantially improve the performance of multi-weather semantic segmentation.

Our contribution is summarized as follows: 
\begin{itemize}
\item We propose a novel computer vision task, which is to model and estimate the co-occurrences of multi-weather through a single image.
\item We propose a novel weather uncertainty modeling scheme. Specifically, it models the representation of each weather as a Gaussian mixture model, which is approximated by prior-posterior learning.
\item We propose a novel \textbf{M}ulti-weather \textbf{E}stimation Trans\textbf{former} (MeFormer) for the task.
It jointly leverages weather-wise token and prior/posterior net to learn weather uncertainty.
\item A new large state dataset named \textbf{M}ulti-w\textbf{e}ather Co-\textbf{p}resence \textbf{e}stimation (MePe) is provided to study the weather uncertainty. It is so far the only dataset that provides both category-level and probability-level ground truth.
\item Large scale experiments are conducted showing the performance of the proposed MeFormer and the MePe dataset.
\item Down stream tasks show large improvement when weather conditions are handled in the proposed continuous and co-occurrences manner.
\end{itemize}

\section{Related Work}
\label{related}

\subsection{Visual Weather Recognition}
In the past decade, the understanding of weather conditions is usually considered as a multi-label classification task \cite{lin2017rscm}.

For the classic machine learning pipelines, Lin \etal \cite{lin2017rscm} designed a region selection strategy and calculated the concurrency to recognize multiple regions that are related to the weather conditions.
In the deep learning era, many fully-supervised multi-label classification methods have been proposed.
They can be categorized into recurrent neural network based \cite{zhao2018cnn} and graph neural network \cite{xie2021graph} based solutions.
More recently, some attention has been paid to multi-weather recognition under the self-supervised settings \cite{xie2022learning,xie2022wcatn}, as large-amount weather annotations from human experts are usually difficult to obtain.

However, none of the existing methods can determine the probability of the weather conditions.

\subsection{Uncertainty Quantification}
\label{sec2.3}
Uncertainty is well studied in machine learning and non-task specific settings \cite{kendall2017uncertainties}.
Given a distribution along with a model, its predictive uncertainty is a combination of data uncertainty and model uncertainty. 
The true posterior distribution is intractable by using naive Bayes’ rule without the approximation.
Thus, either an explicit or implicit variation is usually used for approximation.

Typical solutions include classic ensembles \cite{Gal2016Dropout,Laksh2017Simple}, Bayes neural networks \cite{sengupta2020ensembling,foong2020expressiveness}, and explicit network learning \cite{ciosek2020conservative,kopetzki2021evaluating,hsu2020generalized,Malinin2018Predictive}. Explicit network learning based uncertainty estimation alleviates the computational cost of classic ensembles and Bayes neural networks, and estimates the uncertainty in an end-to-end manner.
An usual way of explicit network learning is to exploit a prior net for explicit approximation \cite{Ji2021Learning,ciosek2020conservative,Malinin2018Predictive}.

\subsection{Uncertainty-aware Vision}
Uncertainty-aware vision received increasingly attention in the past few years.  
Specially, Bertoni et al. \cite{Ber2019monopose} introduced the model uncertainty for more precise pedestrian localization from monocular images. Yang \etal \cite{Yang2021DSUI} proposed a dual-supervised uncertainty inference (DS-UI) framework, which exploits the uncertainty inside the feature representations extracted from the model. Lu et al. \cite{Lu2021geometry} introduced the impact of depth confidence for the object detection task, and designed a Geometry Uncertainty Projection Network. 
Similar uncertainty modeling is also widely used in many other vision tasks such as camouflaged object detection \cite{Yang2021uncertain,Li2021Uncertainty}, domain adaptation \cite{Guan2021UncertaintyA,Wang2021Uncertainty1,Teja2021Uncertainty} and semantic segmentation \cite{Zhou2020Uncertainty2}.

On the other hand, the uncertainty of vision dataset has also been extensively studied. Specially, Peterson \etal \cite{Peter2019human} designed a soft label based CIFAR10 dataset, which is relabelled by multiple human annotations to approximate the data uncertainty. Khan \etal \cite{khan2019striking} models both the class-level and sample-level uncertainty for imbalanced image classification. Hu \etal \cite{Hu2020Uncertainty} improved the performance of zero-shot semantic segmentation by exploiting the uncertainty from the noisy training data. Ji \etal \cite{Ji2021Learning} exploited and modeled the data uncertainty from multiple observations.

However, \textit{to the best of knowledge}, the uncertainty of weather conditions has not been studied so far.


\section{Modeling Multi-weather Uncertainty}
\label{preli}

In this section, we first analyze the multi-weather co-existence and model the formulation from a physical perspective (Sec.~\ref{sec3.1}).
Next, we discuss the problem formulation limitation of existing multi-weather classification methods, and propose our multi-weather formulation, demonstrating the necessity of uncertainty modeling (Sec.~\ref{def3.1}).
Then, we briefly review the uncertainty quantification (Sec.~\ref{sec3.2}) and leverage it to model the weather uncertainty (Sec.~\ref{uncertain3.4}).
Finally, we introduce our proposed prior-posterior learning based weather uncertainty
(Sec.~\ref{sec3.5}).

\subsection{Physics Background of Multi-weather Co-existence}
\label{sec3.1}

\subsubsection{Multi-Weather Observation}

Multiple weather conditions can co-exist in an outdoor scenario.
Take two weather conditions, namely, sunny and rainy, as an example.
As is shown in the first row of Fig.~\ref{observation}, the co-existence between sunny and rainy can a variety of weather appearance as a transitional state.
In some cases, the weather appearance shows stronger sunny than rainy, vice versa.
Similar observation can also be found between rainy and foggy, as shown in the second row of Fig.~\ref{observation}. The intense change of rain and fog is continuous. In some scenes, the appearance of rain can be stronger, but in other scenes, the appearance of fog can be stronger.

To better understand this observation, we turn to the weather formulation from a meteorology perspective.
A feasible path to describe sunny and rainy is through the Moisture Conservation Equation \cite{zhang2004some}. The moisture difference between sunny and rainy is usually obvious. 
Given time $t$, moisture storage $S$, moisture inflow from the 
atmosphere $M_I$, moisture outflow from the 
atmosphere $M_O$, evapotranspiration from ground $E$ and precipitation $P$,
according to the Moisture Conservation Equation \cite{zhang2004some}, the moisture change in the atmosphere can be mathematically expressed as
\begin{equation} 
\label{MCE}
\frac{\partial S}{\partial t} = M_I + E - M_O - P.
\end{equation}

Then, whether the weather condition $s_i$ belongs to sunny $s_{sunny}$ or $s_{rainy}$ is conventionally determined by a binary hard classification:
\begin{equation} 
\label{MeteSunny}
s_{i} =
\begin{cases}
s_{sunny} & \text{if  $S$ \textless$ S_{T}$ }\\
s_{rainy} & \text{if  $S$ \textgreater$ S_{T}$}
\end{cases},
\end{equation} 
where $S_{T}$ denotes the moisture threshold to categorize sunny from rainy.

In fact, for a majority of semantics within the context of computer vision, such hard classification and constellation labels work well. For example, cat and dog can be labeled without ambiguity. Existing weather recognition methods \cite{zhao2018cnn,xie2021graph,xie2022learning,xie2022wcatn} follow such routine. First, an image is embedded into a deterministic feature space. Then, a hard decision is used to make a binary prediction. Hence, existing methods recognize weather in a yes/no manner.
For example, Is Scene-A sunny? Yes. Is Scene-A rainy? No. Is Scene-B foggy? Yes. Is Scene-B rainy. Yes.

However, due to the impact of multiple variants (\eg, $t$, $M_I$, $M_O$, $E$, $P$) and their combinations, the weather condition is a transitional state, and can co-exist with each other. Thus, given different threshold $S_T$, the probability of sunny and rainy can be quite different. A hard classification boundary is less precise to describe the overlap region between sunny and cloudy.

In contrast, our objective is to directly learn the representation of moisture storage $S$ in Eq.~\ref{MCE} and model the transitional state as continuous, which is better to measure the co-existence and probability between sunny and rainy.

\subsubsection{Multi-weather Formulation}

Without loss of generality, here we start to consider the formulation of multiple weather conditions instead of only binary sunny-rainy conditions. In fact, the co-existence can occur between three or more weather conditions, which we denote as $s_1$, $s_2$, $\cdots$, $s_n$.

Notice that, the Moisture Conservation Equation (in Eq.~\ref{MCE}) is applicable not only to sunny and rainy, but also to many other weather conditions. For example, considering the formulation of fog. According to the Dew Point Fog Formulation \cite{wallace2006atmospheric} Equation, the fog density $d$ is determined by the temperature $T$ and a relative form of moisture $\tilde{S}$, given by
\begin{equation} 
\label{fogdensity}
d = \frac{\alpha \cdot T}{\beta + T} + {\rm ln}\tilde{S},
\end{equation}
where $\alpha$ and $\beta$ are constants for the Magnus-Tetens approximation.

Without loss of generality, we extend the moisture storage $S$ in Eq.~\ref{MCE}, and re-define it as a moisture-related parameter $X$ to describe the state of weather conditions in an outdoor scene.

On top of this, we further analyze how the co-existence between multiple weather conditions and impact visual representation learning.
Specifically, given a scene radiance $\mathbf{J}$, an image $\mathbf{I}$ that carries the weather information can be formulated as
\begin{equation} 
\label{weatherfor}
\mathbf{I} = f_I(\mathbf{J}, X),
\end{equation}
where $f_I(\cdot)$ is an image formulation function.
Eq.~\ref{MCE}~and~\ref{weatherfor}~indicate that, the moisture-related parameter $X$ observed from the atmosphere in an image $\mathbf{I}$ is determined by a variety of variables. Different combination of such variables makes the co-existence and varied probabilities of multiple weather conditions a common phenomenon.

\subsection{Multi-weather Co-presence Estimation: Problem Formulation}
\label{def3.1}

In this paper, we propose a novel task in computer vision, \ie, to estimate the multi-weather conditions directly from the physics-related parameter $X$ (in Eq.~\ref{weatherfor}) as a continuous state, instead of the hard classification by the existing methods.

\subsubsection{Formulation of Multi-weather Classification}

We first theoretically analyze the limitation of previous multi-weather classification methods in a machine learning view. 
For these works, assume a feature extractor $g(\cdot)$, parameterized by model parameter $\theta$, extracts the weather-wise embedding $\mathbf{z}_{s_1}$, $\cdots$, $\mathbf{z}_{s_n}$. Then, the confidence score of each weather category can be mathematically computed as
\begin{equation} 
\label{weatherscore}
p(s_{i} | \theta) = \frac{e^{\mathbf{z}_{s_i}}}{\sum_n e^{\mathbf{z}_{s_i}}},
\end{equation}
with a multi-label classification learning objective $\mathcal{L}$, given by
\begin{equation} 
\label{multilabel}
\mathcal{L} = \frac{1}{n} \sum_{i=1}^{n} Y_{s_i} \cdot p(s_{i} | \theta).
\end{equation}

However, it is rather difficult and challenging for such hard classification based methods to distinguish the gradient of each weather condition.
Specifically, the gradient of a weather condition $s_i$ in this scenario is computed as
\begin{equation} 
\begin{aligned}
\label{Lweather}
\frac{\partial \mathcal{L}}{\partial s_i} & = \frac{\partial \mathcal{L}}{\partial f_I} \cdot \frac{\partial f_I}{\partial \mathbf{J}} \cdot \frac{\partial \mathbf{J}}{\partial \mathbf{z}_{s_i}} \cdot \frac{\partial \mathbf{z}_{s_i}}{\partial s_{i}} + \frac{\partial \mathcal{L}}{\partial f_I} \cdot \frac{\partial f_I}{\partial X} \cdot \frac{\partial X}{\partial \mathbf{z}_{s_i}} \cdot \frac{\partial \mathbf{z}_{s_i}}{\partial s_{i}} \\ 
& = \frac{\partial \mathcal{L}}{\partial f_I} \cdot \frac{\partial f_I}{\partial \mathbf{J}} \cdot \frac{\partial \mathbf{J}}{\partial \mathbf{z}_{s_i}} \cdot \frac{\partial \mathbf{z}_{s_i}}{\partial s_{i}} \\ & + \frac{\partial \mathcal{L}}{\partial f_I} \cdot \frac{\partial f_I}{\partial X} \cdot (\frac{\partial X}{\partial M_I} \cdot \frac{\partial M_I}{\partial \mathbf{z}_{s_i}} \cdot \frac{\partial \mathbf{z}_{s_i}}{\partial s_{i}} + \frac{\partial X}{\partial E} \cdot \frac{\partial E}{\partial \mathbf{z}_{s_i}} \cdot \frac{\partial \mathbf{z}_{s_i}}{\partial s_{i}})
\\ & - \frac{\partial \mathcal{L}}{\partial f_I} \cdot \frac{\partial f_I}{\partial X} \cdot (\frac{\partial X}{\partial M_O} \cdot \frac{\partial M_O}{\partial  \mathbf{z}_{s_i}} \cdot \frac{\partial \mathbf{z}_{s_i}}{\partial s_{i}} + \frac{\partial X}{\partial P} \cdot \frac{\partial P}{\partial \mathbf{z}_{s_i}} \cdot \frac{\partial \mathbf{z}_{s_i}}{\partial s_{i}}).
\end{aligned}
\end{equation} 

From a machine learning perspective, Eq.~\ref{Lweather}~is impacted by much more variables than conventional multi-label classification tasks, making the model much more difficult to converge when optimization. Specifically, based on a variety of combinations between $M_I$, $M_O$, $P$ and $E$, scenarios can happen:
\begin{itemize}
\item A same weather condition $s_i$ from different images can have dramatically different variables of $M_I$, $M_O$, $P$ and $E$.
\item Different weather conditions between $s_i$ and $s_j$ can have similar variables of $M_I$, $M_O$, $P$ and $E$.
\item A same weather condition $s_i$ from different images can have dramatically different gradient value $\frac{\partial \mathcal{L}}{\partial s_i}$.
\item Different weather conditions between $s_i$ and $s_j$ can have similar gradient value $\frac{\partial \mathcal{L}}{\partial s_i}$.
\end{itemize}

\begin{figure}[!t]
  \centering
   \includegraphics[width=1.0\linewidth]{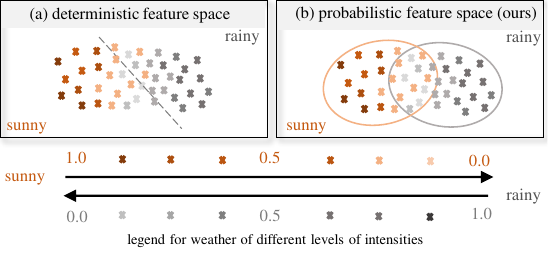}
   \caption{
   Given a binary sunny-rainy example, feature space illustration on: (a) conventional deterministic pipelines; (b) our probabilistic pipeline with uncertainty quantification. The probabilistic feature space is more capable to describe the scenario where multi-weather conditions co-exist.
   }
   \label{weatherform}
\end{figure}

To conclude, from a joint meteorology and machine learning perspective, conventional deterministic multi-class classification models are less precise to perceive the gradients and less discriminative to describe the weather from an image (illustrated in Fig.~\ref{weatherform}a).

\subsubsection{Multi-weather Co-presence Estimation}


Instead of binary and deterministic recognition, this paper defines the problem of estimating weather as a status with probability: Assume there are $n$ weather conditions, denoted as $s_{1}$, $\cdots$, $s_{i}$, $\cdots$, $s_{n}$, we use $p_{s_{i}} \in [0, 1]$ to quantify the probability/confidence of the occurrence, where $p_{s_{i}} = 1$ refers to a condition with high probability. The task of this paper is to estimate the probability of weather conditions from a single image with the awareness of co-existence. Given an image $\mathbf{I}$, a prediction head $h$ estimates the probability $p_{s_{i}}$ of each weather condition $s_{i}$ via the deep feature $\mathbf{X}$ that represents the moisture-related parameter $X$:
\begin{equation}
\mathbf{X} \stackrel{h}{\longrightarrow} \{\hat{p}_{s_{i}}\}, i = 1 \cdots n,
\label{eq:prob_def}
\end{equation}
where $\mathbf{X}$ is learnt by an image encoder $g$ via $\mathbf{X}=g(\mathbf{I})$.

A directly probability prediction instead of the conventional multi-label classification formulation (Eq.~\ref{multilabel}) is more capable to describe the probability of each weather condition. Besides, to better describe the co-existence under different level probabilities, we further introduce the uncertainty quantification on top of Eq.~\ref{eq:prob_def} (illustrated in Fig.~\ref{weatherform}b).

\subsection{RECAP on Uncertainty Quantification}
\label{sec3.2}

Uncertainty can be classified into two categories, that is, \textit{aleatoric uncertainty} (also referred to data uncertainty) and \textit{epistemic uncertainty} (also referred to model uncertainty). 
Given a distribution $P(x,y)$ over the inputted features $x$ and label $y$, the predictive uncertainty of the model $\phi$, which we denote as $P(w|x, D)$ on this finite dataset $D=\{x_{j}, y_{j}\}_{j=1}^{N} \sim P(x,y)$, is a combination of data uncertainty and model uncertainty, given by
\begin{equation} \label{combinationAUEU}
P(w|x, D) = \int \underbrace{P(w|x, \theta)}_{\text{Data}} \underbrace{P(\theta|D)}_{\textit{Model}} d \theta ,
\end{equation}
where $\theta$ and $w$ denotes the parameters for model and uncertainty, respectively. Moreover, $P(w|x, \theta)$ denotes the data uncertainty, and $P(\theta|D)$ denotes the model uncertainty. 

However, the true posterior distribution $P(\theta|D)$ is intractable by using naive Bayes’ rule without the approximation. Thus, either an explicit or implicit variational approximation $q(\theta)$ is needed, given by
\begin{equation} \label{posteriorappro}
P(\theta|D) \approx q(\theta) .
\end{equation}

Typical approximation approaches include classic modeling, explicit network learning and etc.
For \textit{classic modeling} \cite{Gal2016Dropout,Laksh2017Simple}, the uncertainty estimation is usually approximated by
\begin{equation} \label{posteriorapprosample}
P(w|x, D) =  \frac{1}{M} \sum_{i=1}^{M} P(w|x, \theta^{i}),   \theta^{i} \sim q(\theta),
\end{equation}
where $M$ denotes the number of model $\theta$ for approximation, and $\theta^{i}$ is the model parameter in the $i^{th}$ sampling process.

For \textit{explicit network learning approaches}, the variation parameter $q_{\theta}(w)$ is usually used to approximate the posterior distribution $P(w|x, D)$. 
To this end, the Kullback–Leibler (KL) divergence is minimized, given by
\begin{equation} \label{KLmeauresum}
KL(q_{\theta}(w)||P(w|x,D)) =  \sum_{w} q_{\theta}(w) log \frac{q_{\theta}(w)}{P(w|x,D)}.
\end{equation}


\subsection{Modeling Weather Uncertainty}
\label{uncertain3.4}

\subsubsection{Understanding from Feature-level}

After feature embedding, rather than a hard decision (Fig.~\ref{weatherform}a), weather conditions are distributed in certain ranges. The distribution may overlap with other conditions reflecting co-occurrence and ambiguity (Fig.~\ref{weatherform}b).

As our objective is to model the uncertainty and ambiguity of the weather conditions in deep learning pipelines, it is necessary to adapt the explicit network learning based uncertainty paradigm in Eq.~\ref{KLmeauresum}.
To begin with, 
a $M$-dimension Gaussian mixture model \cite{rasmussen1999infinite,kendall2017uncertainties} is defined for each of the weather condition $s_i$:
\begin{equation}
    \mathcal{N}(\boldsymbol{\mu}_{s_{i}}, \boldsymbol{\sigma}_{s_{i}}).
    \label{eq:gmm}
\end{equation}

The core of our learning scheme is a low-dimensional latent space $\mathbb{R}^{M}$ to represent the distribution of each weather condition $s_{i}$.
The uncertainty score $a_{s_i}$ of weather $s_{i}$ is defined as
\begin{equation}
    a_{s_i} \triangleq \mathbb{E}[\boldsymbol{\sigma}_{s_{i}}] = \frac{1}{M}\sum_{m=1}^{M} \sigma_{s_{i}}^{m}.
    \label{unscore}
\end{equation}

\subsubsection{Understanding from Pipeline Level}

\begin{figure}[!t]
  \centering
   \includegraphics[width=1.0\linewidth]{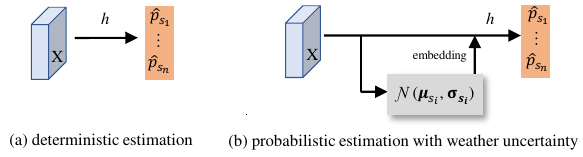}
   \vspace{-0.5cm}
   \caption{Illustration of multi-weather co-presence estimation by (a) existing deep learning paradigm in a deterministic manner; (b) the proposed paradigm with weather uncertainty. $\mathbf{X}$, $p$, $h$ denote the feature representation, probability prediction and prediction head, respectively.  
   }
   \label{learningpipeline}
\end{figure}

As illustrated in Fig.~\ref{learningpipeline}a, for existing deep learning pipelines, multi-weather co-presence estimation (Eq.~\ref{eq:prob_def}) is implemented in a deterministic way. The deep feature $\mathbf{X}$ from feature extractor $g$ is directly mapped to multi-weather probabilities $\{\hat{p}_{s_{i}}\}$ by a prediction head $h$, which usually consists of a MLP followed by a sigmoid activation function. 

In contrast, as illustrated in Fig.~\ref{learningpipeline}b, for the proposed weather uncertainty pipeline, multi-weather co-presence estimation (Eq.~\ref{eq:prob_def}) is implemented in a probabilistic way. The deep feature $\mathbf{X}$ from feature extractor $g$ is additionally modeled with uncertainty (Eq.~\ref{eq:gmm}).
The feature after uncertainty modeling is mapped to multi-weather probabilities $\{\hat{p}_{s_{i}}\}$ by the prediction head $h$. 

\subsection{Prior-posterior Learning for Weather Uncertainty}
\label{sec3.5}

Referring to Fig.~\ref{weatherform}.b and Eq.~\ref{eq:gmm}, a learning strategy is needed such that: on the one hand, a certain weather condition can be located in a certain area of Gaussian; on the other hand, a Gaussian can be located in the area of a certain weather. To this end, for this problem, we propose a prior-posterior learning scheme \cite{ciosek2020conservative,Malinin2018Predictive}.

\begin{figure}[!t]
  \centering
   \includegraphics[width=1.0\linewidth]{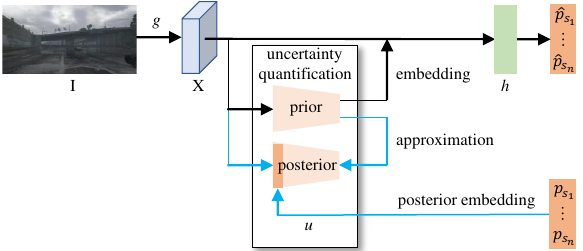}
   \vspace{-0.3cm}
   \caption{The proposed multi-weather co-presence estimation pipeline by modeling the uncertainty of weather features. \textbf{Black} arrows exist in both training \& inference stage, while \textbf{\textcolor{blue}{blue}} arrows only exist in training stage. $\mathbf{I}$, $\mathbf{X}$, $p$, $\hat{p}$ denote the image appearance, feature representation, probability prediction and ground truth, respectively.   
   }
   \label{learningscheme}
\end{figure}

\subsubsection{Prior}
Assume a prior $\mathbf{w}_{prior}$. The prior probability distribution $P(\cdot|\mathbf{X})$ is modelled as an axis-aligned Gaussian with mean $\boldsymbol{\mu}_{prior}(\mathbf{X};\mathbf{w}_{prior})$ and variance $\boldsymbol{\sigma}_{prior}(\mathbf{X};\mathbf{w}_{prior})$. The variant vector $\boldsymbol{z}$ is learnt by $\boldsymbol{z} \sim P(\cdot|\mathbf{X})$, where
\begin{footnotesize}
\begin{equation} \label{priorsample}
P(\cdot|\mathbf{X}) = \mathcal{N}( 
 \boldsymbol{\mu}_{prior}(\mathbf{X};\mathbf{w}_{prior}), diag(\boldsymbol{\sigma}_{prior}(\mathbf{X};\mathbf{w}_{prior}))) .
\end{equation}
\end{footnotesize}

\subsubsection{Posterior}
The posterior probability distribution $Q(\cdot|\mathbf{X}, \mathbf{P})$, parameterized by $\mathbf{w}_{post}$, maps the weather representation to a position $\boldsymbol{\mu}_{post}(\mathbf{X}, \mathbf{P}; \mathbf{w}_{post})$ with uncertainty $\boldsymbol{\sigma}_{post}(\mathbf{X}, \mathbf{P}; \mathbf{w}_{post})$. The variant $\boldsymbol{z}$ is learnt via $\boldsymbol{z} \sim Q(\cdot|\mathbf{X}, \mathbf{P})$, where
\begin{scriptsize}
\begin{equation} \label{posteriorsample}
\begin{aligned}
Q(\cdot|\mathbf{X}, \mathbf{P}) = \mathcal{N}(\mu_{post}(\mathbf{X}, \mathbf{P}; \mathbf{w}_{post}), diag(\sigma_{post}(\mathbf{X}, \mathbf{P}; \mathbf{w}_{post}))).
\end{aligned}
\end{equation}
\end{scriptsize}

\subsubsection{Kullback-Leibler Loss} 
The Kullback-Leibler loss \cite{hall1987kullback} is utilized to minimize the difference between the posterior distribution and prior distribution as follows:
\begin{footnotesize}
\begin{equation} \label{KLmeaureforprior}
\begin{aligned}
& \mathcal{L}_{KL}(Q(\cdot|\mathbf{X}, \mathbf{P})||P(\cdot|\mathbf{X}))= \\ & \mathbb{E}_{z \sim Q(\cdot|\mathbf{X}, \mathbf{P})} \rm{log} Q(\cdot|\mathbf{X}, \mathbf{P}) [\rm{log} Q(\cdot|\mathbf{X}, \mathbf{P}) - \rm{log} P(\cdot|\mathbf{X})].
\end{aligned}
\end{equation}
\end{footnotesize}

\begin{figure*}[!t]
  \centering
   \includegraphics[width=1.0\linewidth]{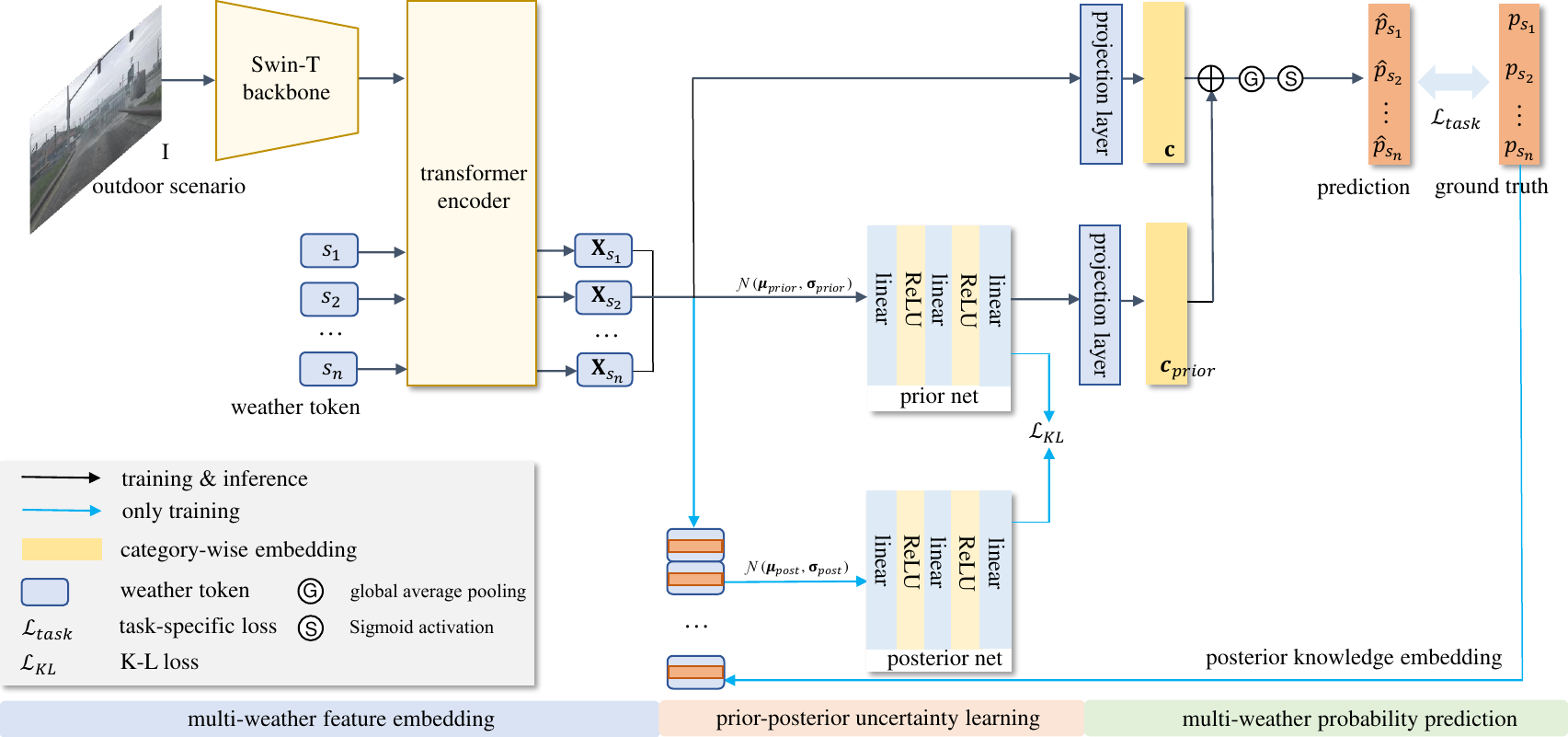}
   \caption{Illustration of the proposed \textbf{M}ulti-weather Co-presence \textbf{E}stimation Trans\textbf{former} (MeFormer). It consists of three key components, namely, multi-weather feature embedding (MFE) (Sec.~\ref{sec4.2}), prior-posterior uncertainty learning (PUL) (Sec.~\ref{sec4.3}), and multi-weather probability prediction (MPP) (Sec.~\ref{sec4.4}). MeFormer is versatile to both conventional multi-label weather classification task and the proposed multi-weather co-presence estimation task, depending on the type of available ground truth.}
   \label{framework}
\end{figure*}

\subsubsection{Pipeline}
As shown in Fig.~\ref{learningscheme}, our uncertainty-aware multi-weather co-presence estimation pipeline plugs in a prior-posterior uncertainty learning (PUL) component $u(\cdot)$ between the feature embedding $g(\cdot)$ and the prediction head $h(\cdot)$. 

\section{Multi-weather Co-presence Estimation Transformer (MeFormer)}
\label{method}
We propose a \textbf{M}ulti-weather \textbf{E}stimation Trans\textbf{former} (MeFormer), shown in Fig.~\ref{framework}. It contains three major modules: the multi-weather feature embedding (MFE, in Sec.~\ref{sec4.2}), the prior-posterior uncertainty learning (PUL, in Sec.~\ref{sec4.3}) and the multi-weather probability prediction (MPP, in Sec.~\ref{sec4.4}). Sec.~\ref{sec4.5} will describe the loss functions.

\subsection{Multi-weather Feature Embedding}
\label{sec4.2}
The multi-weather feature embedding (MFE) module is the first part of the pipeline. A set of $n$ \textit{weather token} is embedded in a self-attention mechanism to the Swim-Transformer \cite{liu2021swin}, where $n$ is the number of weather conditions.
The feature embedding is given by:
\begin{equation}
    \mathbf{X} = [\mathbf{X}_{s_{1}}, \mathbf{X}_{s_{2}}, \cdots, \mathbf{X}_{s_{n}}], 
\label{eq:token}
\end{equation}
where $\mathbf{X}_{s_{i}} \in \mathbb{R}^{h \cdot w}$ is the $i^{th}$ feature learnt from the transformer. $h$ and $w$ are the spatial dimension of the transformer's output.

Then, we stack $\mathbf{X}$ with the original output of Swim-Transformer $\mathbf{x} = [\mathbf{x}_1, \cdots, \mathbf{x}_c]$, denoted by $\mathbf{H} = [\mathbf{x}, \mathbf{X}]$. $\mathbf{x}$ is reshaped to $hw \times c$ so that it has the same height as $\mathbf{X}$. $c$ is the number of channels from the original output.

Then, for any two elements in $\mathbf{H}$, denoted by $\mathbf{h}_{l}$ and $\mathbf{h}_{m}$, the importance $a_{l,m}$ between $\mathbf{h}_{l}$ and $\mathbf{h}_{m}$ is measured by the transformer encoder as follows:
\begin{equation} \label{selfattQK}
\begin{aligned}
a_{l,m} = \rm{softmax} ((\boldmath{W}_{\textit{Q}} \mathbf{h}_{\textit{l}})^\mathrm{T} (\mathbf{W}_{\textit{K}} \mathbf{h}_{\textit{m}})/ \sqrt{c} ),
\end{aligned}
\end{equation}
\begin{equation} \label{selfattV}
\begin{aligned}
\mathbf{\overline{h}}_{l} = \sum_{m=1}^{w \times h} a_{l,m} \mathbf{W}_{V} \mathbf{h}_{m},
\end{aligned}
\end{equation}
\begin{equation} \label{selfembed}
\begin{aligned}
\mathbf{h}_{l'} = \mathbf{W}_{2} \rm{ReLU}(\mathbf{W}_{1} \mathbf{\overline{h}}_{\textit{l}} + \mathbf{b}_{1})+ \mathbf{b}_{2},
\end{aligned}
\end{equation}
where $\mathbf{W}_{Q}$, $\mathbf{W}_{K}$ and $\mathbf{W}_{V}$ denote the query, key and value matrix in self-attention respectively. $\rm{Softmax}$ and $\rm{ReLU}$ denote the Softmax normalization and ReLU activation function respectively. $\mathbf{\overline{h}}_{l}$ denotes the intermediate embedding. Also, $\mathbf{W}_{1}$, $\mathbf{b}_{1}$ and $\mathbf{W}_{2}$, $\mathbf{b}_{2}$ are the weight and bias matrix of two linear layers respectively. The feature value of $\mathbf{X}_{s_{i}}$ is the representation of weather category $s_{i}$.

\subsection{Prior-posterior Uncertainty Learning}
\label{sec4.3}

Following Eq.~\ref{priorsample}, the prior network is implemented as a three-layer conditional variational auto-encoder (cVAE) \cite{sohn2015learning}, which learns $\mathbf{z}_{prior}$ from $\mathbf{X}$. It carries the $\boldsymbol{\mu}_{prior}$ and $\boldsymbol{\sigma}_{prior}$ from each weather condition. Specifically, the first, second and third linear layer have 64, 32 and 16 dimensions respectively, and the first and second linear layer are both followed by a ReLU activation function. Here, the low-dimensional latent space size $M$ is equal to 16 (in Eq.~\ref{eq:gmm}).

Similarly, following Eq.~\ref{posteriorsample}, the posterior network has the same cVAE architecture as the prior net, which learns the posterior variant vector $\mathbf{z}_{post}$, represented by $\boldsymbol{\mu}_{post}$ and $\boldsymbol{\sigma}_{post}$. The difference between the prior and posterior network is that the ground truth is embedded into the input of the posterior network as posterior knowledge. Specifically, the ground truth $p_{s_{i}}$ is added into each channel of $w \times h$ dimensions. 

Finally, the K-L loss $\mathcal{L}_{KL}$ (Eq.~\ref{KLmeaureforprior}) is used to minimize the difference between $\mathbf{z}_{prior}$ and $\mathbf{z}_{post}$, so that the posterior distribution is approximated by the prior distribution.

\subsection{Multi-weather Co-presence Prediction}
\label{sec4.4}
For the Multi-weather Probability Prediction, a category-level representation $\mathbf{c}$ is generated by a projection layer parameterized by $\mathbf{W}_{3}$, $\mathbf{b}_{3}$ with $\mathbf{X}$ as input.
Similarly, another presentation $\mathbf{c}_{prior}$ is generated by another projection layer parameterized by $\mathbf{W}_{4}$, $\mathbf{b}_{4}$ with $\mathbf{z}_{prior}$ as input.

\begin{equation} \label{cenhanced}
\begin{aligned}
\setlength{\abovedisplayskip}{1pt}
\mathbf{c} = \mathbf{W}_{3} \mathbf{X} + \mathbf{b}_{3},
\setlength{\belowdisplayskip}{1pt}
\end{aligned}
\end{equation}
\begin{equation} \label{cprior}
\begin{aligned}
\setlength{\abovedisplayskip}{1pt}
\mathbf{c}_{prior} = \mathbf{W}_{4} \mathbf{z}_{prior}+ \mathbf{b}_{4}.
\setlength{\belowdisplayskip}{1pt}
\end{aligned}
\end{equation}

Both category-wise representations are fed into the regression head and the final output of the weather estimation is calculated by:
\begin{equation} \label{pfinal}
\begin{aligned}
\setlength{\abovedisplayskip}{1pt}
\mathbf{\hat{P}} = \rm{Sigmoid} (\rm{GAP}(\mathbf{c}_{prior}+ \mathbf{c})),
\setlength{\belowdisplayskip}{1pt}
\end{aligned}
\end{equation}
where $\mathbf{\hat{P}}=[\hat{p}_{s_{1}}, \hat{p}_{s_{2}}, \cdots, \hat{p}_{s_{n}}]$. Also, $\rm{Sigmoid}$ and $\rm{GAP}$ denote the Sigmoid activation function and the global average pooling. 

\subsection{Loss Functions}
\label{sec4.5}

The proposed MeFormer can be supervised by both binary-level label and probability-level label, depending on the type of available ground truth.
Besides, the proposed MeFormer can also be adapted to other weather-related tasks.

Take the co-presence estimation task as an example, the MSE loss $\mathcal{L}_{MSE}$ is used between ground truth $p_{s_{i}}$ and prediction $\hat{p}_{s_{i}}$:
\begin{equation} \label{Lmse}
\begin{aligned}
\setlength{\abovedisplayskip}{1pt}
\mathcal{L}_{MSE} = \frac{1}{n} \sum_{i=1}^{n}||\hat{p}
_{s_{i}}- p_{s_{i}}||^{2}.
\setlength{\belowdisplayskip}{1pt}
\end{aligned}
\end{equation}

The total loss function $\mathcal{L}$ is a linear combination of the regression loss $\mathcal{L}_{MSE}$ and the uncertainty estimation loss $\mathcal{L}_{KL}$ with a weight $\lambda$, presented as: 
\begin{equation} 
\begin{aligned}
\setlength{\abovedisplayskip}{1pt}
\mathcal{L} = \mathcal{L}_{MSE} + \lambda \mathcal{L}_{KL},
\setlength{\belowdisplayskip}{1pt}
\label{loss}
\end{aligned}
\end{equation}
with weight $\lambda$.


\section{MePe Dataset}
\label{dataset}
A new dataset is collected: \textbf{M}ulti-w\textbf{e}ather Co-\textbf{P}resence \textbf{e}stimation (MePe) dataset.
MePe has four key advantages: (1) a large number of samples with a large diversity of weather conditions and combinations, (2) the weather conditions are controlled by realistic simulations such that the probability of each weather can be modeled properly, (3) it includes a variety of outdoor scenes and time of the day, (4) it is so far the only multi-weather dataset that provides both binary-label and probability-label ground truth.

\subsection{Scientific Definition of Weather Categories}
\label{def}
\subsubsection{Meteorology Definition}

As a prerequisite, we investigate the weather definitions from weather forecasting  \cite{ailliot2015stochastic}, weather science \cite{piotrowicz2020selection,lolis2020use} and meteorology \cite{ahrens2015meteorology,wallace2006atmospheric}, and summarize 14 weather categories for our studies. The detailed definitions and citations are provided in the first and second column of Table~\ref{weatherdefreal}.

These categories cover most common weather conditions. Some of these categories are quite similar, and some of them tend to co-occur with several other categories. Such design provides enough diversity and great challenge to model and validate the weather uncertainty. In addition, our categorization also meets the demand of emerging real-world applications that need fine-grained weather categorization. For example: 

\textbf{1) Categorization between blizzard and snow, rain and drizzle.} 
First of all, in meteorology \cite{ahrens2015meteorology}, there is rigid definition and distinction between common snow and blizzard (Table~\ref{weatherdefreal}). 
Also, this difference is particularly important for some safety-crucial visual applications like autonomous driving \cite{diaz2022ithaca365,sakaridis2021acdc}. Small snow grains are safe to continue the driving, while the blizzard with snow cover is more necessary to give the decision basis like speed limitation or even brake.

\textbf{2) Categorization between fog and haze.} 
First, in meteorology \cite{ahrens2015meteorology}, there is rigid definition and distinction between common rain and drizzle. Then, recent work also infers the haze and fog as different categories for guidance of human activities \cite{han2021joint}). 

\textbf{3) Categorization between clear and extra-sunny.}
First, in meteorology \cite{ahrens2015meteorology}, there is rigid distinction between clear and extra-sunny. Clear can occur in both day or night time. Extra-sunny is in day-time without any cloud in the sky, and is usually accompanied with the warning of sunburn \cite{boldeman2001tanning}. In spirit of the human activity guidance \cite{han2021joint}, extra-sunny is also included in our weather categorization. 

\begin{table*}[!t]
\begin{center}
\caption{Definition of 14 weather types from: (1) (Column 1\& 2) meteorology \cite{ahrens2015meteorology} and weather forecasting; (2) (Column 3\& 4) GTAV engine.}
\label{weatherdefreal}
\resizebox{\linewidth}{!}{
\begin{tabular}{c|c|c|c}
\hline
\multicolumn{2}{c|}{meteorology and weather forecasting} & \multicolumn{2}{c}{GTAV engine} \\
\hline
Category & Definition & Category & Definition \\
\hline
sunny/clear	& \makecell[c]{there are clouds above the horizon. \\ The sky is clear, or no cloud in the night} & clear & \makecell[c]{sunny; no cloud in the sky; \\ either in day or night} \\
\hline
clearing & \makecell[c]{an intermediate weather condition \\ that changes from rainy to clear} & clearing & \makecell[c]{rain is stopping, \\ and it is becoming sunny} \\
\hline
clouds & \makecell[c]{significant amount of clouds is covering \\ the sky (at least half the sky)} & cloudy & \makecell[c]{clouds in the sky; can't see \\the sun; shadows on the ground} \\
\hline
overcast & \makecell[c]{the sky is completely covered by a cloud blanket, 
\\ and becoming dimmer/completely dark} & overcast & 
\makecell[c]{clouds obscures almost \\all the sky} \\
\hline
rain & \makecell[c]{condensed moisture of the atmosphere \\ above falling under the form of liquid droplets} & rainy & \makecell[c]{clouds obscures almost \\ all the sky} \\
\hline
drizzle	& \makecell[c]{similar to rain, but the droplets are very small \\ and hardly noticeable with the naked eye.} & light rain & \makecell[c]{raindrop is smaller, the sky \\ is brighter than rain scenario} \\
\hline
snow & \makecell[c]{snow is atmospheric water that \\ froze and fell to the ground} & snow & \makecell[c]{ice crystals shows up; \\ the sky gets dark} \\
\hline
stormy & \makecell[c]{characterized by lightning, and produced by \\ the largest and tallest clouds that can spawn} & thunder & \makecell[c]{a rain-bearing cloud \\ is producing lightning} \\
\hline
haze & \makecell[c]{aggregation in atmosphere of widely-dispersed particles. \\ an opalescent air appearance that subdues colors} & smog & \makecell[c]{the air is \\community-wide polluted} \\
\hline
neutral & \makecell[c]{the weather condition with cloud in the sky. \\ a 50 percent chance of seeing above-average rainfall} & neutral & \makecell[c]{there is about 50\% chance \\ that it is going to rain} \\
\hline
extrasunny & \makecell[c]{day time that has more sunlight than clear, \\ with no cloud in the sky and the risk of sunburn} & extra sunny & \makecell[c]{more sunlight than clear \\ scenario; no cloud in the sky} \\
\hline
fog & \makecell[c]{cloud at ground level, which raises ambient humidity \\ to its maximum, and considerably decreases visibility} & foggy & \makecell[c]{fog shows up \\ in the environment} \\
\hline
frozen & \makecell[c]{the surface air temperature is expected \\ below 32°F for a climatologically significant period of time} & frozen & \makecell[c]{after snow or rain, \\ it is frozen outside} \\
\hline
blizzard & \makecell[c]{a severe snowstorm characterized by \\ strong sustained winds and low visibility} & blizzard & \makecell[c]{ice crystal is larger; \\ the snow cover is more} \\
\hline
\end{tabular}
}
\end{center} 
\end{table*}

\subsubsection{Definition Correspondence in GTAV Engine}
Based on our investigated 14-type weather categorization, we pick 14 weather categories in Grand Theft Auto V (GTAV) that correspond to the definition in our categorization.
The 14 selected weather conditions in GTAV that correspond to the 14-type weather categories are \textit{foggy}, \textit{snow}, \textit{smog}, \textit{neutral}, \textit{rainy}, \textit{blizzard}, \textit{clear}, \textit{light rain}, \textit{cloudy}, \textit{overcast}, \textit{clearing}, \textit{frozen}, \textit{thunder} and \textit{extra sunny}. Their definition in GTAV and correspondence to the real-world weather \cite{ahrens2015meteorology} is summarized in the third and forth column of Table~\ref{weatherdefreal}.
It can be seen that:
all these 14 weather categories selected from GTAV engine can correspond to the real-world meteorology weather categories \cite{ahrens2015meteorology}.

\subsection{Dataset Overview}
\label{sec5.1}

\renewcommand{\arraystretch}{1.0} 
\begin{table*}[!t]
\caption{Summary of existing multi-weather datasets. \#Samples: sample number; \#Cat.: weather category number; gt: ground truth. Existing multi-weather datasets only provide category-level annotation without probability ground truth.}
\label{datasetinfo}
\resizebox{\linewidth}{!}{
\begin{tabular}{c|cc|cccc|cc}
\hline
Dataset & \# Samples & \# Cat. & cat. gt & pro. gt & fine-grained cat. & high resolution & image size & train/test split\\
\hline
Transient Attribute Dataset \cite{laffont2014transient} & 8,571 & 7 & $\checkmark$ & $\times$ & $\times$ & $\times$ & 256$\times$256 & 80\% / 20\% \\
Weather Attribute Dataset \cite{lin2017rscm} & 10,000 & 6 & $\checkmark$ & $\times$ & $\times$ & $\times$ & 256$\times$256 & 50\% / 50\% \\
Multi-weather Recognition Dataset \cite{zhao2018cnn} & 10,000 & 5 & $\checkmark$ & $\times$ & $\times$ & $\times$ & 256$\times$256 & 80\% / 20\% \\
\hline
MePe (Ours) & 16,078 & 14 & $\checkmark$ & $\checkmark$ & $\checkmark$ & $\checkmark$ & 2048$\times$1024 & 80\% / 20\% \\
\hline
\end{tabular}
}
\end{table*}

\begin{figure*}[!t]
  \centering
   \includegraphics[width=1.0\linewidth]{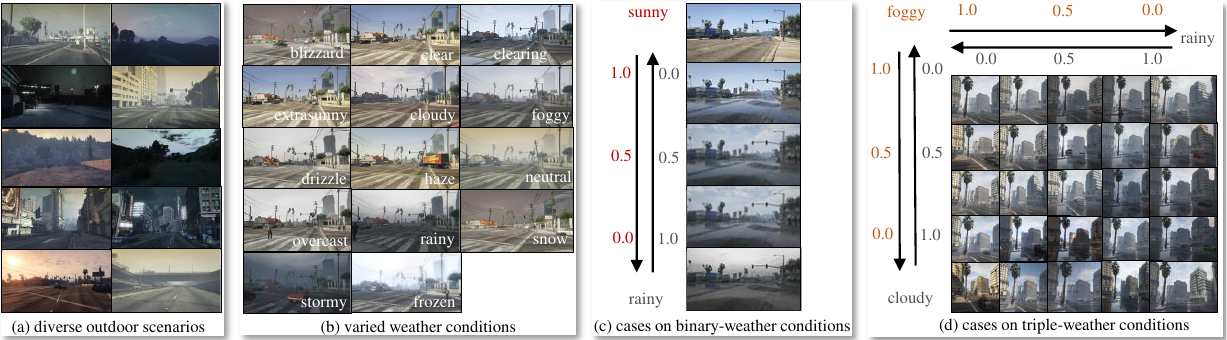}
   \caption{Overview of the multi-weather co-presence estimation (MePe) Dataset. (a): diverse \& complicated outdoor views. (b): 14 different weather conditions. (c) \& (d): blending examples of binary and triple weather conditions. 
   }
   \label{dataset}
\end{figure*}

MePe contains 14 single weather conditions, namely, \textit{foggy}, \textit{snow}, \textit{smog}, \textit{neutral}, \textit{rainy}, \textit{blizzard}, \textit{clear}, \textit{snowlight}, \textit{cloudy}, \textit{overcast}, \textit{clearing}, \textit{frozen}, \textit{thunder} and \textit{extra sunny}. 
MePe has a total of 16,078 samples and each of them has single or multiple dominant weather conditions.
The training set contains 80\% samples and the testing set contains 20\%. Fig.~\ref{dataset} shows a number of samples. 

Table.~\ref{datasetinfo} shows a comparison with existing datasets \cite{zhao2018cnn,lin2017rscm}. MePe is the only dataset that can estimate multi-weather probabilities. MePe also has more training samples and weather conditions.

\subsection{Dataset Generation}
\textbf{Weather rendering.} 
As shown in Fig.~\ref{weatherform}a, it is difficult to manually label the ambiguities between weather conditions. To this end, we use the computer generated imagery (CGI) rendering to steer the probability of each weather condition. The Grand Theft Auto V (GTAV) is used as the rendering platform. It has been widely utilized in computer vision for generating large-scale and photo-realistic datasets \cite{richter2016playing,sekkat2020omniscape}.

GTAV allows to precisely control weather conditions $s_i$ as well as extra parameters such as probability, illumination, blurring, fog density, cloud density, and raindrop scale.
To generate a single weather, all the parameters are kept default, and the specific weather is assigned a dominant value so that the single weather is generated without uncertainty. Fig.~\ref{dataset}b shows the visual effect by overlaying a single weather condition to the scene.
To generate multiple weather conditions, we consider the widely-used Richardson-Wright Stochastic weather model \cite{wilks1999weather} in the meteorology field, which uses linear fusion between two weather categories.
Specifically, we overlay the single effects in a linear way.
Fig.~\ref{dataset}.c and .d show overlays of binary and triple weather conditions under various probabilities.

\textbf{Probability approximation.}
As mentioned above, given that multi-weather conditions are generated using linear blending (Fig.~\ref{dataset}.c and .d), a linear interpolation is used to quantize the probability of the weather. 
For an image $\mathbf{I}$, the weight $a_{s_{i}}$ is approximated as the probability $p_{s_{i}}$ of the weather condition $s_{i}$, \ie,
$p_{s_{i}} \approx a_{s_{i}}$,
where $a_{s_{i}} \in [0, 1]$. 

\textbf{Error analysis.} 
The error $\sigma_{a_{s_{i}}}^{2}$ caused by the weight approximation on the multi-weather co-presence estimation learning objective (Eq.~\ref{Lmse}) can be calculated via $\sigma_{L}^{2}=(\sigma_{a_{s_{i}}}^{2})^{2}$. It indicates that a 10\% error from the GTAV approximation on the ground truth only has a 1\% error impact on the model learning process, which is acceptable and practical.


\subsection{Evaluation Protocols}

\subsubsection{Track 1: Multi-weather co-presence estimation}

For co-presence estimation \cite{zhao2017learning, zhao2017approximating, Peter2019human}, four metrics are used, namely, the sum of squared difference (SSD), the Kullback-Leibler divergence (KL), coefficient of determination ($R^{2}$) and cross-entropy (CE).
\begin{equation} \label{metric1}
\setlength{\abovedisplayskip}{1pt}
SSD = \sum_{i=1}^{n}|p_{s_{i}} -\hat{p}_{s_{i}}|^{2},
\setlength{\belowdisplayskip}{1pt}
\end{equation}
\begin{equation} \label{metric2}
\setlength{\abovedisplayskip}{1pt}
KL = \sum_{i=1}^{n} p_{s_{i}} log (\frac{p_{s_{i}}}{\hat{p}_{s_{i}}}),
\setlength{\belowdisplayskip}{1pt}
\end{equation}
\begin{equation} \label{metric4}
\setlength{\abovedisplayskip}{1pt}
R^{2} = 1 - \frac{\sum_{i=1}^{n}(p_{s_{i}}-\hat{p_{s_{i}}})^{2}}{\sum_{i=1}^{n}(p_{s_{i}}-\frac{1}{n}\sum_{i=1}^{n}p_{s_{i}})^{2}}
\setlength{\belowdisplayskip}{1pt}
\end{equation}
\begin{equation} \label{metric4}
\setlength{\abovedisplayskip}{1pt}
CE = -  \sum_{i=1}^{n} p_{s_{i}} log(\hat{p}_{s_{i}}).
\setlength{\belowdisplayskip}{1pt}
\end{equation}

$SSD$, and $KL$ describe the distance between the prediction and ground truth, $R^{2}$ measures the correlation between the prediction and ground truth, and $CE$ measures the uncertainty of prediction.


\subsubsection{Track 2: Multi-weather Classification}

Following the prior Transient Attribute Dataset \cite{laffont2014transient} and Multi-weather Recognition Dataset \cite{zhao2018cnn}, both per-weather and per-sample evaluation are involved.
For per-weather evaluation, three metrics, namely,
average precision (AP), average recall (AR), and average F1 score (AF1), are used.
For per-sample evaluation, three metrics, namely, overall precision (OP), overall recall (OR), and overall F1 score (OF1), are used. 
Here the average/overall F1 score is harmonic mean of the average-/overall- precision and recall, respectively.


\section{Experiments and Analysis}
\label{experiment}

\subsection{Results on Multi-weather Co-presence Estimation}
\label{exproest}

\begin{table*}[!t]
\scriptsize
\begin{center}
\caption{Per-category weather estimation performance of the deterministic estimation (without MFE and PUL modules, denoted as w/o unc.) and the proposed MeFormer (denoted as ours) on the MePe dataset. Percentages in \textcolor{red}{\textbf{red}} indicate improvement while in \textcolor{blue}{\textbf{blue}} indicate degradation.}
\label{perweather}
\resizebox{\linewidth}{!}{
\begin{tabular}{cc|cccccccccccccc|c}
\hline
Metric & Model & \rotatebox{90}{blizzard} & \rotatebox{90}{clear} & \rotatebox{90}{clearing} & \rotatebox{90}{cloudy} & \rotatebox{90}{extrasunny} & \rotatebox{90}{foggy} & \rotatebox{90}{neutral} & \rotatebox{90}{overcast} & \rotatebox{90}{rain} & \rotatebox{90}{smog}& \rotatebox{90}{snow}  & \rotatebox{90}{snowlight}  & \rotatebox{90}{thunder}  & \rotatebox{90}{frozen} & \rotatebox{90}{all} \\
\hline
\multirow{3}*{SSD ($ \times 10^{-3}$) $\downarrow$} & w/o unc. & 1.493 & 2.829 & 2.411 & 2.213 & 2.144 & 1.750 & 1.899 & 2.184 & 1.553 & 2.066 & 1.929 & 2.156 & 1.308 & 2.498 & 2.037 \\
~ & ours & 1.045 & 1.818 & 1.179 & 1.420 & 1.732 & 1.332 & 2.144 & 1.366 & 1.058 & 1.122 & 1.393 & 1.190 & 0.572 & 1.417 & 1.340 \\
~ & ~ & \textcolor{red}{-30.0\%} & \textcolor{red}{-35.8\%} & \textcolor{red}{-51.1\%} & \textcolor{red}{-35.8\%} & \textcolor{red}{-19.2\%} & \textcolor{red}{-23.9\%} & 
\textcolor{blue}{+12.9\%} & \textcolor{red}{-37.5\%} & \textcolor{red}{-31.9\%} & \textcolor{red}{-45.7\%} & \textcolor{red}{-27.8\%} & \textcolor{red}{-44.8\%} & \textcolor{red}{-56.3\%} & \textcolor{red}{-43.3\%} & \textcolor{red}{-34.2\%} \\
\hline
\multirow{3}*{KL ($ \times 10^{-3}$) $\downarrow$} & w/o unc. & 3.486 & 6.726 & 5.778 & 5.323 & 5.090 & 4.233 & 4.598 & 5.194 & 3.834 & 4.955 & 4.653 & 5.284 & 3.072 & 5.870 & 4.874 \\
~ & ours & 2.470 & 4.414 & 2.828 & 3.417 & 4.080 & 3.205 & 5.253 & 3.241 & 2.569 & 2.624 & 3.385 & 2.791 & 1.275 & 3.324 & 3.205 \\
~ & ~ & \textcolor{red}{-29.2\%} & \textcolor{red}{-34.4\%} & \textcolor{red}{-51.1\%} & \textcolor{red}{-35.8\%} & \textcolor{red}{-19.8\%} & \textcolor{red}{-24.3\%} & 
\textcolor{blue}{+14.2\%} & \textcolor{red}{-37.6\%} & \textcolor{red}{-33.0\%} & \textcolor{red}{-47.0\%} & \textcolor{red}{-27.3\%} & \textcolor{red}{-47.2\%} & \textcolor{red}{-58.5\%} & \textcolor{red}{-43.4\%} & \textcolor{red}{-34.2\%} \\
\hline
\multirow{3}*{$R^{2}$ $\uparrow$} & w/o unc. & 0.829 & 0.675 & 0.743 & 0.749 & 0.703 & 0.800 & 0.779 & 0.743 & 0.815 & 0.762 & 0.766 & 0.781 & 0.814 & 0.774 & 0.762 \\
~ & ours & 0.883 & 0.793 & 0.865 & 0.841 & 0.752 & 0.844 & 0.746 & 0.845 & 0.876 & 0.874 & 0.828 & 0.878 & 0.929 & 0.868 & 0.843 \\
~ & ~ & \textcolor{red}{6.5\%} & \textcolor{red}{17.5\%} & \textcolor{red}{16.5\%} & \textcolor{red}{12.2\%} & \textcolor{red}{7.0\%} &  \textcolor{red}{5.5\%} & \textcolor{blue}{-4.2\%} & \textcolor{red}{13.7\%} & \textcolor{red}{7.6\%} & \textcolor{red}{14.6\%} & \textcolor{red}{8.1\%} & \textcolor{red}{12.5\%} & \textcolor{red}{14.1\%} & \textcolor{red}{12.1\%} & \textcolor{red}{+10.6\%} \\
\hline
\multirow{2}*{CE $\downarrow$} & w/o unc. & 1.361 & 1.395 & 1.329 
 & 1.369 & 1.383 & 1.373 & 1.327 & 1.384 & 1.327 & 1.374 & 1.397 & 1.326 & 1.329 & 1.323 & 1.375 \\
~ & ours & 1.125 & 1.127 & 1.126 & 1.125 & 1.127 & 1.126  & 1.128 
& 1.126 & 1.126 & 1.125 & 1.128 & 1.123  & 1.127 & 1.120 & 1.124 \\
~ & ~ & \textcolor{red}{-17.3\%} & \textcolor{red}{-19.2\%}  & \textcolor{red}{-15.3\%}  & \textcolor{red}{-17.8\%}  & \textcolor{red}{-18.5\%}  & \textcolor{red}{-18.0\%} & \textcolor{red}{-15.0\%}  & \textcolor{red}{-18.6\%}  & \textcolor{red}{-15.2\%}  & \textcolor{red}{-18.1\%}  & \textcolor{red}{-19.2\%}  & \textcolor{red}{-15.3\%}  & \textcolor{red}{-15.2\%}  & \textcolor{red}{-15.3\%} & \textcolor{red}{-18.3\%} \\
\hline
\end{tabular}  }
\end{center} 
\end{table*}

\subsubsection{Implementation Details}
The Swim-Transformer is pretrained on ImageNet \cite{liu2021swin} and the rest is initialized randomly.
The Adam optimizer is used with an initial learning rate of $2\times10^{-4}$.
The weight decay is set to $1\times10^{-4}$, and the dropout rate is set to 0.1. The training terminates after 50 epochs. The batch size is set 2 per GPU. $\lambda$ in Eq.~\ref{loss} is set to $1\times10^{-5}$.

All the experiments are conducted on the proposed MePe dataset, which is so far the only dataset that provides the precise probability-level weather ground truth.

\subsubsection{Weather-wise performance}
Given that the task is novel, results are only compared by the proposed method and the baseline, which removes the prior-posterior uncertainty learning (without MFE and PUL modules, denoted as w/o unc.). 
A breakdown analysis on 14 weather conditions is shown in Table.~\ref{perweather}. 
It can be derived that with uncertainty modeling, the performance is improved on all weathers when using $CE$, and is significantly improved on 13 out of 14 weather conditions when using $SSD$, $KL$ and $R^2$. Especially, the improvement is most significant on \textit{smog}, \textit{snowlight}, \textit{thunder} and \textit{frozen}. \textit{Neutral} is the only condition where the performance is not always improved. This is because the uncertainty aware module decreased the probability on \textit{neutral}.

To sum up, the proposed method innovatively formulates the multi-weather uncertainty from a physical perspective, and incorporates it into an explicit learning pipeline.
Its feature representation is much stronger than existing deterministic pipelines to describe the multi-weather uncertainty. 

\subsubsection{Performance on the number of mixed weather}
In Table~\ref{sanity}, we further evaluate the capability to estimate multiple weather conditions according to how complex the scene is. It is shown that the uncertainty modeling constantly and significantly improves the performance on all metrics and on all levels: from two dominant conditions to more than four conditions.

As sanity check, images are used with a dominant single weather to demonstrate that the proposed method also works well on simple cases. From the first row of Table.~\ref{sanity}, the proposed MeFormer works adequately for simple weather condition.

\begin{table}[!t]
\begin{center}
\caption{Performance regarding complexity of weather conditions. ours: the proposed MeFormer. w/o unc.: the MeFormer with MFE and PUL modules removed, for deterministic estimation. Experiments conducted on the test set of MePe dataset. Percentages in \textcolor{red}{\textbf{red}} indicates better improvement while in \textcolor{blue}{\textbf{blue}} indicates worse degradation.}
\label{sanity}
\resizebox{\linewidth}{!}{
\begin{tabular}{cc|cccc}
\hline
\#category & Model & SSD ($ \times 10^{-3}$) $\downarrow$ & KL ($ \times 10^{-3}$) $\downarrow$ & $R^{2}$ $\uparrow$ & CE $\downarrow$ \\
\hline
\multirow{3}*{1} &  w/o unc. & 2.455	& 5.307 & 0.690 & 1.363 \\
~ & ours & 1.412 & 3.304 & 0.816 & 1.131 \\
~ & ~ & \textcolor{red}{-42.5\%} & \textcolor{red}{-37.7\%} & \textcolor{red}{+18.3\%} & \textcolor{red}{-17.0\%} \\
\hline
\multirow{3}*{2} &  w/o unc. & 1.172	& 4.126 & 0.776 & 1.306 \\
~ & ours & 0.959 & 2.238 & 0.869 & 1.120 \\
~ & ~ & \textcolor{red}{-18.2\%} & \textcolor{red}{-45.8\%} & \textcolor{red}{+12.0\%} & \textcolor{red}{-14.2\%} \\
\hline
\multirow{3}*{3} &  w/o unc. & 2.186 & 5.247 & 0.751 & 1.347 \\
~ & ours & 1.462 & 3.512 & 0.835 & 1.126 \\
~ & ~ & \textcolor{red}{-33.1\%}  & \textcolor{red}{-33.1\%} & \textcolor{red}{+11.2\%} & \textcolor{red}{-16.4\%} \\
\hline
\multirow{3}*{4} &  w/o unc. & 2.098 & 5.046	& 0.765 & 1.386 \\
~ & ours & 1.518 &	3.650 & 0.831 & 1.125 \\
~ & ~  & \textcolor{red}{-27.6\%}  & \textcolor{red}{-27.7\%} & \textcolor{red}{+8.6\%} & \textcolor{red}{-18.8\%} \\
\hline
\multirow{3}*{\textgreater 4} &  w/o unc. & 1.831 & 4.413 &	0.801 & 1.395 \\
~ & ours & 1.417 & 3.463 & 0.843 & 1.125\\
~ & ~ & \textcolor{red}{-22.6\%}  & \textcolor{red}{-21.5\%} & \textcolor{red}{+5.2\%} & \textcolor{red}{-19.4\%} \\
\hline
\multirow{3}*{All} &  w/o unc. & 2.037 & 4.874 & 0.762 & 1.375 \\
~ & ours & 1.340 & 3.205 & 0.843 & 1.124 \\
~ & ~ & \textcolor{red}{-34.2\%}  & \textcolor{red}{-34.2\%} & \textcolor{red}{+10.6\%} & \textcolor{red}{-18.3\%} \\
\hline
\end{tabular}  }
\end{center} 
\end{table}

\subsubsection{Qualitative performance} 
\begin{figure*}
  \centering
  \includegraphics[width=1.0\linewidth]{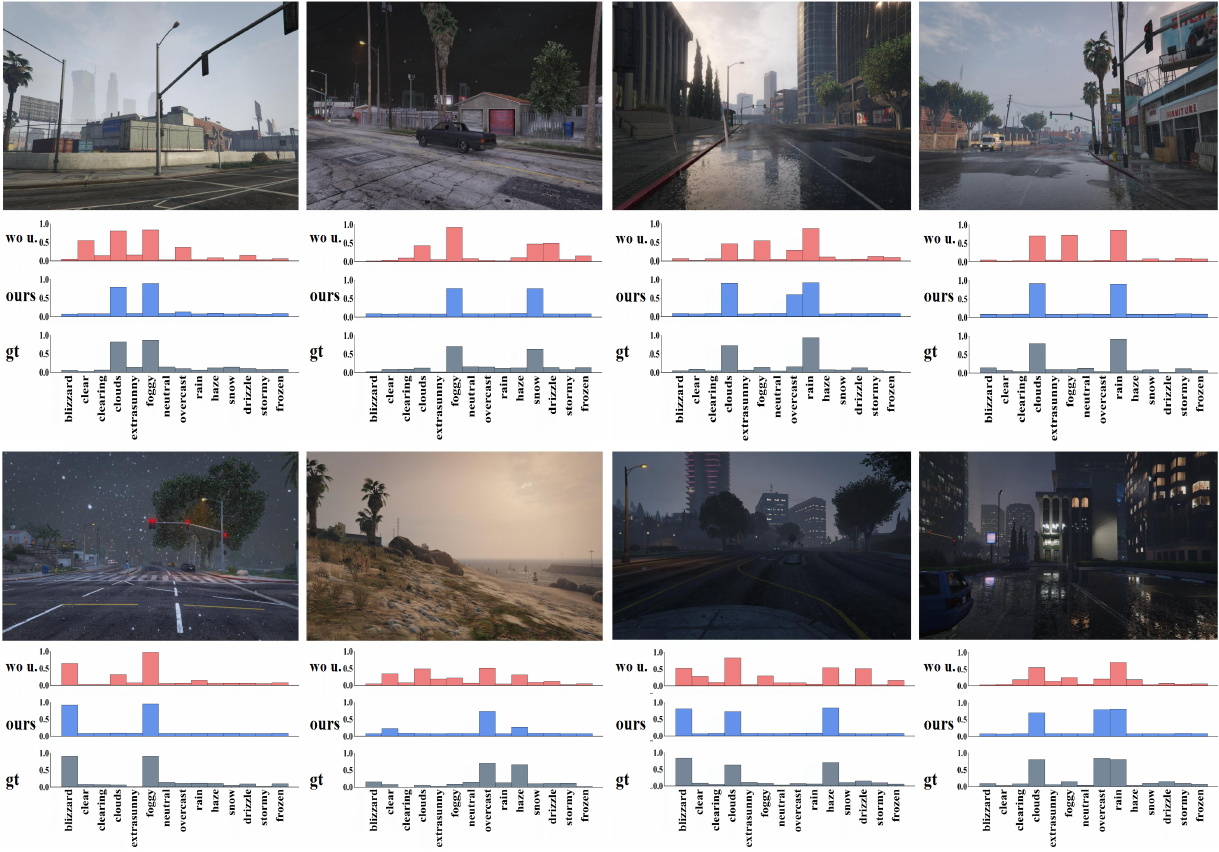}
   \captionof{figure}{Visual results on MePe dataset. w/o unc.: conventional deterministic regression pipelines; ours: the proposed MeFormer with uncertainty quantification.}
   \label{morevis}
\end{figure*}

Fig.~\ref{morevis}~provides a number of visualized multi-weather co-presence estimation results with a variety of weather and co-occurrence cases. Compared to the deterministic estimation pipeline (w/o unc.), the proposed MeFormer provides a significantly better estimation prediction. 

\subsection{Results on Multi-weather Classification}

\subsubsection{Implementation Details}
All the hyper-parameter settings and architecture designs directly follow the experiments on multi-weather co-presence estimation (in Sec.~\ref{exproest}).
The only difference is that, the predicted multi-weather probability is supervised by 0-1 binary labels, and the multi-label cross-entropy loss is adapted. For evaluation, the output is changed to a binary value (\ie, 1 or 0) for each weather condition. 

The experiments are conducted on
Multi-weather Recognition Dataset \cite{zhao2018cnn} and the proposed MePe dataset (with binary label supervision).
Some methods are not open source. Therefore, results are cited from \cite{zhao2018cnn,xie2021graph} accordingly. 
Six evaluation metrics are used, namely, average precision (AP), average recall (AR), and average F1 score (AF1), overall precision (OP), overall recall (OR), and overall F1 score (OF1). 

\subsubsection{Compare with State-of-the-art}

Table~\ref{multirecog}~compares the performance of MeFormer with \cite{zhao2018cnn,xie2021graph,lanchantin2021general,zhu2021residual,zhao2021transformer}. 
On Multi-weather Recognition Dataset, it shows state-of-the-art on the AP, AF1 and OR metrics. 
For the rest AR, OP and OF1 metrics, its performance is also very close to the state-of-the-art. 
On the proposed MePe dataset, it achieves the best performance on all these six metrics.

\begin{table*}[!t]
\begin{center}
\caption{Comparison on the multi-weather recognition task on Multi-weather Classification Dataset \cite{zhao2018cnn} and the proposed MePe dataset. }
\label{multirecog}
\resizebox{\linewidth}{!}{
\begin{tabular}{c|cccccc|cccccc}
\hline
\multirow{2}{*}{method} & \multicolumn{6}{c|}{Multi-weather Recognition Dataset \cite{zhao2018cnn}} & \multicolumn{6}{c}{MePe Dataset} \\
\cline{2-13}
~ & AP & AR & AF1 & OP & OR & OF1 & AP & AR & AF1 & OP & OR & OF1 \\
\hline
Att-ConvLSTM \cite{zhao2018cnn} & 87.2 & 87.0 & 87.1 & 92.6 & 89.5 & 91.4 & 85.5 & 77.9 & 81.5 & 83.2 & 79.8 & 81.5 \\
GCN-A \cite{xie2021graph} & 87.0 & 86.7 & 86.8 & 88.2 & 89.1 & 88.7 & 85.0 & 78.2 & 81.5 & 84.2 & 79.8 & 81.9 \\
C-Trans \cite{lanchantin2021general} & 87.8 & 87.8 & 87.8 & 89.1 & 89.3 & 89.2 & 84.7 & 79.5 & 82.0 & 83.2 & 81.3 & 82.3 \\
CSRA \cite{zhu2021residual} & 79.9 & 78.5 & 79.2 & 81.0 & 80.9 & 81.0 & 90.0 & 83.1 & 86.4 & 88.5 & 84.8 & 86.6 \\
TDRG \cite{zhao2021transformer} & 88.7 & 85.6 & 87.5 & 87.8 & 83.1 & 85.0 & 89.8 & 83.0 & 86.3 & 88.3 & 84.7 & 86.5 \\
\hline
Ours & \textbf{89.6} & 86.5 & \textbf{88.0} & 88.7 & \textbf{89.6} & 86.5 & \textbf{90.7} & \textbf{85.0} & \textbf{88.7} & \textbf{89.4} & \textbf{85.9} & \textbf{87.6}  \\
\hline
\end{tabular}
}
\end{center} 
\end{table*}

\subsubsection{Performance on Category-wise Classification}

We further compare the category-wise weather classification performance between the proposed MeFormer and the state-of-the-art on further conduct experiments on the proposed MePe dataset. 
Following its evaluation protocol, the accuracy of each weather category is reported.

Table~\ref{permultirecog}~reports the results.
It achieves the best classification performance on ten out of fourteen weather categories.

\begin{table*}[!t]
\begin{center}
\caption{Per-Weather Classification Comparison on the proposed MePe dataset under the multi-label classification setting. Per-category accuracy is reported. }
\label{permultirecog}
\begin{tabular}{c|cccccccccccccc}
\hline
Method & \rotatebox{90}{blizzard} & \rotatebox{90}{clear} & \rotatebox{90}{clearing} & \rotatebox{90}{cloudy} & \rotatebox{90}{extrasunny} & \rotatebox{90}{foggy} & \rotatebox{90}{neutral} & \rotatebox{90}{overcast} & \rotatebox{90}{rain} & \rotatebox{90}{smog}& \rotatebox{90}{snow}  & \rotatebox{90}{snowlight}  & \rotatebox{90}{thunder}  & \rotatebox{90}{frozen}  \\
\hline
Att-ConvLSTM \cite{zhao2018cnn} & 95.6 & 68.1 & 56.3 & 81.6 & 72.5 & 86.9 & 82.3 & 70.8 & 90.0 & 75.5 & 79.7 & 63.5 & 76.0 & 55.2 \\
GCN-A \cite{xie2021graph} & 96.1 & 74.7 & 70.6 & 66.1 & 70.2 & 87.5 & 84.4 & 76.9 & 83.8 & \textbf{87.3} & 75.0 & 49.5 & \textbf{93.8} & 45.8 \\
C-Trans \cite{lanchantin2021general} & 96.9 & 74.3 & 74.5 & 75.8 & 71.5 & 89.2 & 85.9 & 65.0 & 92.4 & 86.4 & 87.0 & 54.2 & 87.5 & 52.1  \\
CSRA \cite{zhu2021residual} & 97.9 & 75.1 & 75.8 & 80.7 & 73.5 & \textbf{91.2} & 87.0 & 76.0 & 95.2 & 85.9 & 89.8 & \textbf{71.9} & 82.3 & 58.3 \\
TDRG \cite{zhao2021transformer} & 98.1 & 75.9 & 76.0 & 81.0 & 77.1 & 89.9 & 87.5 & 78.0 & 95.5 & 85.2 & 87.2 & 63.5 & 83.3 & 59.4 \\
\hline
Ours & \textbf{98.2} & \textbf{79.6} & \textbf{77.9} & \textbf{81.5} & \textbf{77.3} & 90.0 & \textbf{91.7} & \textbf{79.4} & \textbf{95.5} & 86.5 & \textbf{90.6} & 68.2 & 89.6 & \textbf{64.6} \\
\hline
\end{tabular}
\end{center} 
\end{table*}

\subsection{Results on Multi-weather Semantic Segmentation}

\subsubsection{Implementation Details}

Our uncertainty-aware weather estimation is applied on the ACDC dataset \cite{sakaridis2021acdc}, which is a commonly-use dataset for segmentation with adverse weather conditions.
The proposed MeFormer shares the image encoder with the original segmentation model. The MFE and PUL modules are attached in a parallel manner. The learnt weather prior is embedded after the image encoder by channel-wise broadcast. The four weather categories in ACDC are converted into binary label to supervise the posterior net in the PUL module.

Following the evaluation protocol\cite{sakaridis2021acdc}, we run each experiment setting by three independent repetitions.

\subsubsection{Results}

Results are reported in Table~\ref{ADseg}.
By using uncertainty-aware weather estimation as a prior, the segmentation results on DeepLab-V3 \cite{chen2017rethinking}, SegFormer \cite{xie2021segformer} and Mask2Former \cite{cheng2021per} are improved by 0.8\%, 0.6\% and 0.9\% mIoU, respectively. 
Nevertheless, the segmentation results show a slight decline by 0.3\% mIoU on HRNet \cite{wang2020deep}. 
The performance decline on HRNet may be explained by its less discriminative representation, which is a light-weight model.

Overall, a clear segmentation improvement is observed when leveraging the weather prior. It provides a potential solution to improve the robustness of outdoor scene segmentation under adverse weather conditions.

\begin{table}[!t]
\begin{center}
\caption{Performance on multi-weather semantic segmentation on the ACDC dataset \cite{sakaridis2021acdc}. The proposed MeFormer is stacked to the Mask2Former (denoted as Me+Mask2Former) to convey the weather prior for segmentation. In \textcolor{red}{\textbf{red}} indicates better improvement while in \textcolor{blue}{\textbf{blue}} indicates worse degradation.}
\label{ADseg}
\resizebox{\linewidth}{!}{
\begin{tabular}{c|c|ccc}
\hline
Method & Backbone & w.o. ours & w. ours & $\Delta$ \\
\hline
DeepLabV3 \cite{chen2017rethinking} & ResNet-101 & 70.0 & \textbf{70.8} & \textcolor{red}{$\uparrow$ 0.8} \\
HRNet \cite{wang2020deep} & HRw48 & \textbf{75.0} & 74.7 & \textcolor{blue}{$\downarrow$ 0.3} \\
SegFormer \cite{xie2021segformer} & MiT-B3 & 74.4 & 75.0 & \textcolor{red}{$\uparrow$ 0.6} \\
Mask2Former \cite{cheng2021per} & Swin-B & 76.2 & \textbf{77.1} & \textcolor{red}{$\uparrow$ 0.9} \\
\hline
\end{tabular}
}
\end{center} 
\end{table}

\subsection{Ablation Studies}
\subsubsection{Feature space}
\begin{figure}
  \centering
  \begin{subfigure}{0.48\linewidth}
    \centering
    \includegraphics[width=1.0\linewidth]{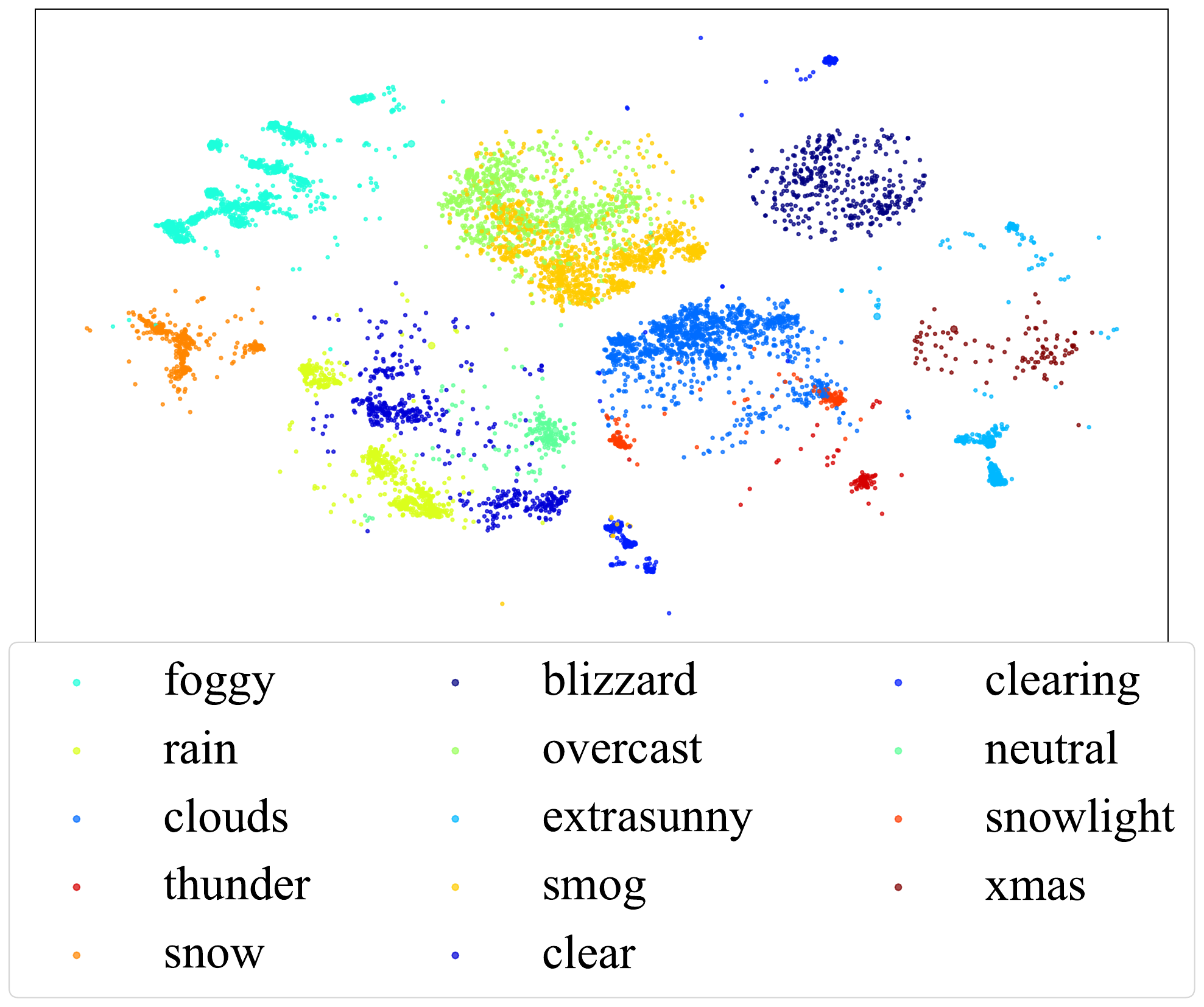}
    \caption{w/o unc.}
    \label{baseline}
  \end{subfigure}
  \hfill
  \begin{subfigure}{0.48\linewidth}
    \centering
    \includegraphics[width=1.0\linewidth]{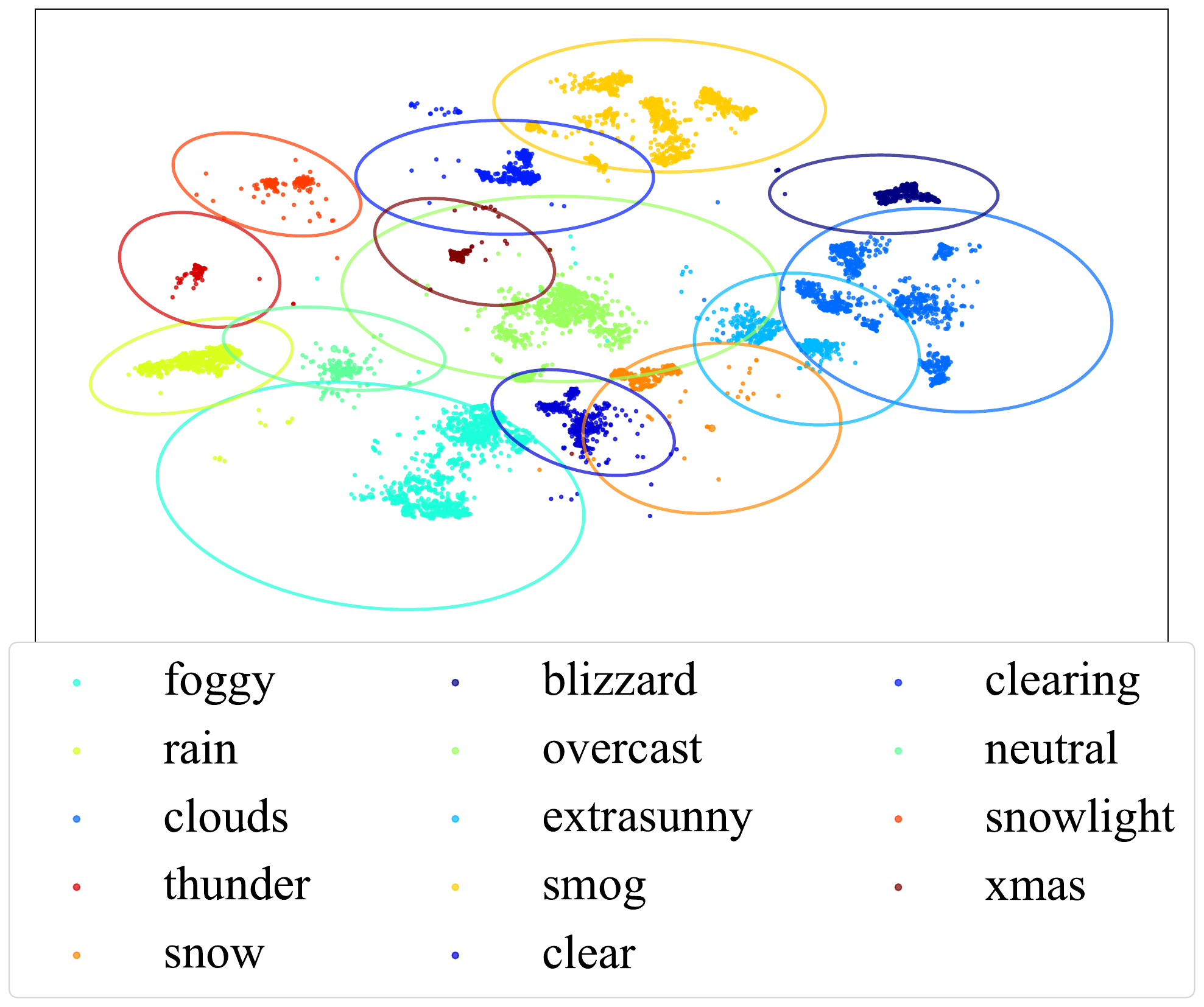}
    \caption{ours}
    \label{ours}
  \end{subfigure}
  \caption{T-sne visualization of the features from the last layer of the (a) deterministic estimation framework (without uncertainty, denoted as w/o unc.), and (b) the proposed framework (ours). By using our framework, the representation is modeled as probabilistic and is more discriminative.}
  \label{tsne}
\end{figure}

Fig.~\ref{tsne} visualizes the feature space (t-sne) of MeFormer without and with the prior-posterior uncertainty learning. It is shown that the uncertainty-aware weather learning scheme not only models the uncertainty but also yields more clustered weather characteristics.

We further study the impact of the dimension of feature spaces (in Eq.~\ref{eq:gmm}). Table~\ref{impactspace} shows that the dimension of the Gaussian mixture module influences the performance at a certain level. A low dimension Gaussian may be insufficient to model the uncertainty while for higher dimensions, the Gaussian may over-fit.

\begin{table}[!t]\scriptsize
\caption{Performance impact of the uncertainty latent space ($M$ in Eq.~\ref{eq:gmm}) on the proposed MeFormer.}
\label{impactspace}
\begin{center}
\resizebox{\linewidth}{!}{
\begin{tabular}{c|cccc}
\hline
size & SSD ($ \times 10^{-3}$) $\downarrow$ & KL ($ \times 10^{-3}$) $\downarrow$ & $R^{2}$ $\uparrow$ & CE $\downarrow$ \\
\hline
4 & 1.500 & 3.590 & 0.826 & 1.126 \\
8 & 1.395 & 3.341 & 0.835 & 1.126 \\
\textbf{16} & \textbf{1.340} & \textbf{3.205} & \textbf{0.843} & \textbf{1.124} \\
24 & 1.356 & 3.247 & 0.842 & 1.126 \\
32 & 1.630 & 3.293  & 0.811 & 1.126 \\
\hline
\end{tabular}
}
\end{center} 
\end{table}

\subsubsection{Performance of each component}
Table~\ref{ablation} provides the effectiveness of each module on the four criterias. All proposed modules provide a consistent contribution to the overall performance.

\begin{table}[!t]
\caption{Impact of each component in the proposed MeFormer. Experiments are conducted on the MePe benchmark. MFE: Multi-weather feature embedding component, PUL: prior-posterior uncertainty learning component.}
\label{ablation}
\begin{center}
\resizebox{\linewidth}{!}{
\begin{tabular}{ccc|cccc}
\hline
backbone & MFE & PUL & SSD ($\times 10^{-3}$) $\downarrow$ & KL ($\times 10^{-3}$) $\downarrow$ & $R^{2}$ $\uparrow$  & CE $\downarrow$  \\
\hline
$\checkmark$ & - & - & 2.037 & 4.874 & 0.762 & 1.375 \\
$\checkmark$ & $\checkmark$ & - & 1.695 & 4.095 & 0.803 & 1.268 \\
$\checkmark$ & - & $\checkmark$ & 1.551 & 3.871 & 0.817 & 1.127 \\
$\checkmark$ & $\checkmark$ & $\checkmark$ & \textbf{1.340} & \textbf{3.205} & \textbf{0.843} & \textbf{1.124} \\
\hline
\end{tabular}
}
\end{center} 
\end{table}

\begin{table}[!t]\scriptsize
\caption{Robustness of the proposed MeFormer when using different loss functions.}
\label{impactloss}
\begin{center}
\resizebox{\linewidth}{!}{
\begin{tabular}{c|cccc}
\hline
loss & SSD ($\times 10^{-3}$) $\downarrow$ & KL  ($\times 10^{-3}$) $\downarrow$ & $R^{2}$ $\uparrow$ & CE $\downarrow$ \\
\hline
$l$-1 loss & 1.395 & 3.332 & 0.836 & 1.127 \\
smooth $l$-1 loss  & 1.366 & 3.268 & 0.841 & 1.126 \\
$l$-2 loss & 1.340 & 3.205 & 0.843 & 1.124 \\
\hline
\end{tabular}
}
\end{center} 
\end{table}

\subsubsection{Choice of regression loss}
While the MeFormer uses $l$-2 loss as a default, it also works robustly on other losses such as $l$-1 and smooth $l$-1. Table~\ref{impactloss} shows the results.

\subsubsection{Influence of Hyper-parameter $\lambda$}
Table~\ref{impactlambda}~reports the performance of the proposed MeFormer when the hyper-parameter $\lambda$ varies on different scales, which balances the impact of co-presence estimation and uncertainty modeling. Generally speaking, the performance of our MeFormer is not much sensitive to the change of $\lambda$ when it is of a large scale. However, when its scale becomes too small, the benefits of uncertainty is negatively impacted and the performance declines.

\begin{table}[!t]\scriptsize
\caption{Performance impact of the hyper-parameter $\lambda$ on the proposed MeFormer.}
\label{impactlambda}
\begin{center}
\resizebox{\linewidth}{!}{
\begin{tabular}{c|cccc}
\hline
$\lambda$ & SSD ($\times 10^{-3}$) $\downarrow$ & KL ($\times 10^{-3}$) $\downarrow$  & $R^{2}$ $\uparrow$ & CE $\downarrow$ \\
\hline
1 $\times 10^{-3}$ & 1.342 & 3.212 & 0.843 & 1.126 \\
1 $\times 10^{-4}$ & 1.346 & 3.223 & 0.842 & 1.126 \\
1 $\times 10^{-5}$ & 1.340 & 3.205 & 0.843 & 1.124 \\
1 $\times 10^{-6}$ & 1.372 & 3.296 & 0.841 & 1.126 \\
1 $\times 10^{-7}$ & 1.420 & 3.467 & 0.837 & 1.128 \\
\hline
\end{tabular}
}
\end{center} 
\end{table}

\subsection{Generalization}
\subsubsection{Performance on real images}
\begin{figure}[!t]
  \centering
   \includegraphics[width=1.0\linewidth]{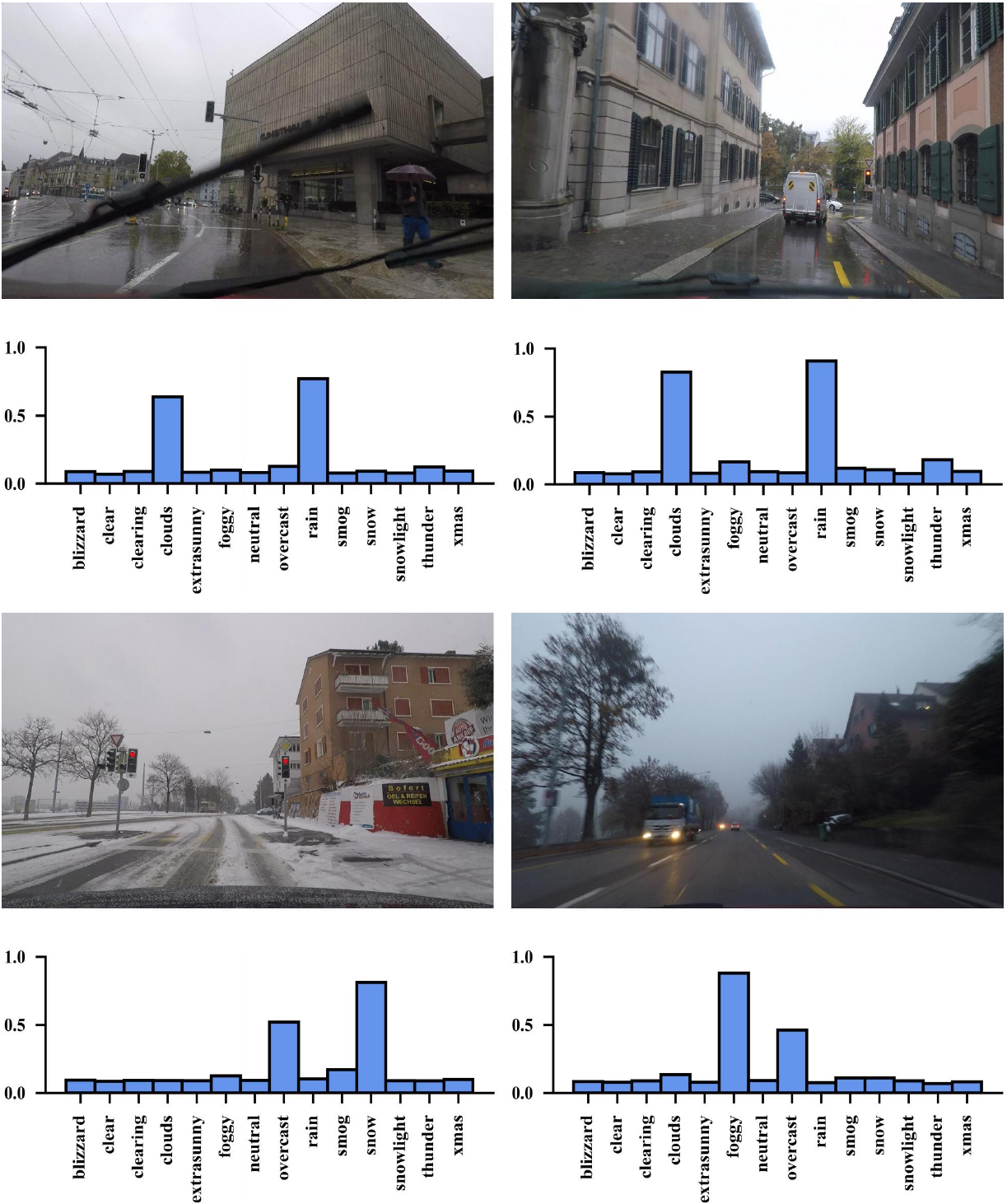}
   \caption{Multi-weather co-presence estimation results on real-world unlabeled outdoor scenes.  
   }
   \label{real}
\end{figure}

We show qualitative results on real images where the ground truth is not available. The pre-trained MeFormer on MePe dataset is used for prediction. Fig.~\ref{real} shows the results demonstrating good generalization of both MeFormer and MePe datasets.

\subsubsection{Failure Cases}
\begin{figure*}[!t]
  \centering  \includegraphics[width=1.0\linewidth,height=0.35\linewidth]{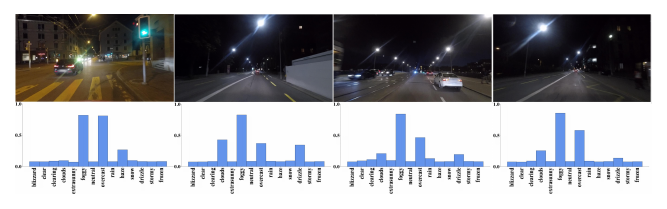}
   \caption{Some failure predictions on unlabeled real-world images from the ACDC dataset \cite{sakaridis2021acdc}. 
   }
   \label{ACDCpredfail}
\end{figure*}

Some failure prediction cases are provided in Fig.~\ref{ACDCpredfail} (with samples from ACDC \cite{sakaridis2021acdc}).

Based on these observations, it can be seen that most wrong inference of the proposed MeFormer can be attributed to the light scattering in the night urban scenarios. In the urban streets there are so many light sources in the night. The light from these light sources usually scatters in the dark sky and dark regions, and there is serve appearance from this phenomenon. Consequently, such appearance, during the inference stage, is misleading for the trained model, and it is more likely to be estimated as the foggy weather condition, which shares similar appearance of the light scattering in terms of the color and shape. 
However, this failure case is not a special issue for the proposed MeFormer. Instead, they are rather common for a variety of methods and tasks related to autonomous driving. 


\section{Conclusion}
\label{conclu}
In this paper, a novel task on multi-weather co-presence estimation is introduced. A novel uncertainty-aware multi-weather learning scheme is proposed by adapting the prior-posterior framework. The scheme is designed in an end-to-end model named MeFormer. A new large-scale multi-weather dataset MePe is provided. Large scale experiments demonstrate the state-of-the-art performance on both the proposed novel task and previous multi-label weather recognition task. Application on improving performance of multi-weather semantic segmentation is also demonstrated.
For future work, we consider further studying the modeling of weather uncertainty and to explore more applications.

\textbf{Limitations.} 
While the probability of each weather can be quantified by simulation, it is infeasible to 
generate ground truth for real-world weather scenarios. We leave this as a research question for the future. 
However, the proposed MeFormer shows generalization capabilities on real-world scenarios when only pre-trained on the synthetic MePe dataset. 

\ifCLASSOPTIONcaptionsoff
  \newpage
\fi



%


\bibliographystyle{IEEEtran}
\bibliography{reference.bib}

%

\begin{IEEEbiography}[{\includegraphics[width=1in,height=1.25in,clip,keepaspectratio]{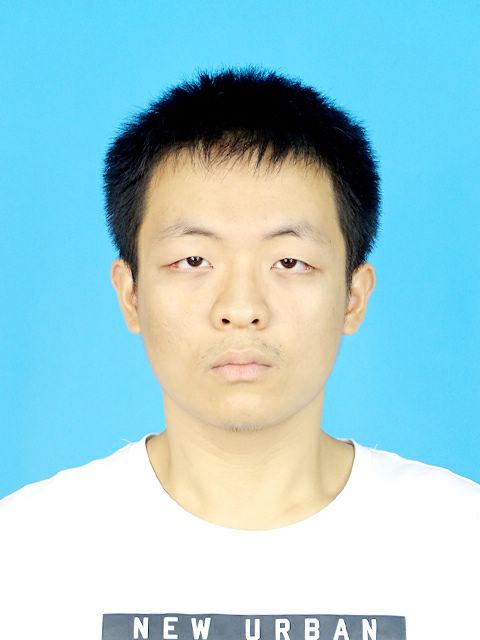}}]{Qi Bi} (Student Member, IEEE) received the B.S. and M.S. degrees from Wuhan University, Wuhan, China, in 2017 and 2020, respectively. He is currently pursuing the Ph.D. degree with the Computer Vision Research Group, University of Amsterdam, Amsterdam, The Netherlands. His research interests include image understanding, robust vision in bad weather and domain generalization. 
He was awarded as an Outstanding Reviewer of CVPR 2023.
His works were shortlisted for CVPR 2021 Best
Paper Candidates and MICCAI 2021 Travel Award.
\end{IEEEbiography}

\begin{IEEEbiography}[{\includegraphics[width=1in,height=1.25in,clip,keepaspectratio]{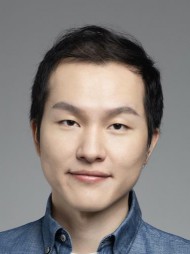}}]{Shaodi You} (Senior Member, IEEE) is an Assistant Professor at University of Amsterdam (UvA), Netherlands. He works in the Computer Vision Research Group in the Institute of Informatics, Faculty of Science. He receives his Ph.D. and M.E. degrees from The University of Tokyo, Japan in 2015 and 2012 and his bachelor's degree from Tsinghua University, P. R. China in 2009. He is best known for low level computer vision especially adversarial weather condition. He serves regularly as area chairs of CVPR, Neurips and ECCV. He is an associate editor of IJCV.
\end{IEEEbiography}

\begin{IEEEbiography}[{\includegraphics[width=1.0in,height=1.25in,clip,keepaspectratio]{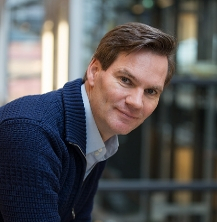}}]{Theo Gevers} (Member, IEEE) is a professor of computer vision at the University of Amsterdam. He is director of the Computer Vision Lab and co-director of the Atlas Lab (UvA-TomTom) and Delta Lab (UvA-Bosch) in Amsterdam. His research area is artificial intelligence with the focus on computer vision and deep learning, and in particular image processing, 3D (object) understanding and human-behavior analysis with industrial and societal applications. He is the co-founder of 3DUniversum, Scanm B.V., and Sightcorp. He has published over 250+ papers and 3 books. He is organizer, general chair, and program committee member of a various number of conferences, and an invited speaker at major conferences.
\end{IEEEbiography}

\end{document}